\documentclass[10pt,hidelinks,journal]{IEEEtai}  % ,final

\usepackage{acronym}
\usepackage[ruled,vlined,linesnumbered]{algorithm2e}
\usepackage{amsmath}
\usepackage{amssymb}
\usepackage{graphicx}
\usepackage{ifdraft}
\usepackage{multirow}
\usepackage{listings}
\usepackage{color}
\usepackage{xspace}
\usepackage{siunitx}
\usepackage{times}
\usepackage{paralist}
\usepackage[bookmarks]{hyperref}
\usepackage{bookmark}
\usepackage[capitalise,noabbrev]{cleveref}
\usepackage{hypernat}
\usepackage{mdframed}
\usepackage{tabulary}
\usepackage{tabularx}
\usepackage{titlesec}
\usepackage{longtable}
\usepackage[font=small,labelfont=bf]{caption}
\usepackage{comment}

\let\labelindent\relax
\usepackage[shortlabels]{enumitem}
\setitemize{noitemsep,topsep=0pt,parsep=0pt,partopsep=0pt}
\setenumerate{noitemsep,topsep=0pt,parsep=0pt,partopsep=0pt}
\setlist{nosep}

\titlespacing*{\section}{0pt}{0.2em}{0.1em}
\titlespacing*{\subsection}{0pt}{0.2em}{0.1em}

\crefname{paragraph}{Section}{Sections}

\DeclareMathOperator*{\argmax}{arg\,max}

\date{\today}

\begin{document}
\bstctlcite{IEEEexample:BSTcontrol}

\newcommand\titletext{Criticality and Safety Margins \\for Reinforcement Learning}
\newcommand\titletextoneline{Criticality and Safety Margins for Reinforcement Learning}
\title{\titletext}
\author{%
Alexander Grushin, Walt Woods, Alvaro Velasquez, and Simon Khan
\thanks{This material is based upon work supported by the Air Force Research Laboratory (AFRL) under Contract No. FA8750-22-C-1002.  This paper was PA approved under case number AFRL-2023-5871.\\
SBIR Data Rights\\
Contract No.: FA8750-22-C-1002\\
Contractor Name: Galois, Inc.\\
Address: 421 SW Sixth Ave., Suite 300, Portland, OR 97204\\
Expiration of SBIR Data Rights Period: 01/18/2042\\
The Government's rights to use, modify, reproduce, release, perform, display, or disclose technical data or computer software marked with this legend are restricted during the period shown as provided in paragraph (b)(4) of the Rights in Other Than Commercial Technical Data and Computer Software - Small Business Innovation Research (SBIR) Program clause contained in the above identified contract. No restrictions apply after the expiration date shown above. Any reproduction of technical data, computer software, or portions thereof marked with this legend must also reproduce the markings.}
\thanks{This work has been submitted to the IEEE for possible publication. Copyright may be transferred without notice, after which this version may no longer be accessible.}
\thanks{A. Grushin was with Galois, Inc. E-mail: \texttt{agrushin@galois.com}}
\thanks{W. Woods was with Galois, Inc. E-mail: \texttt{woodswalben@gmail.com}}
\thanks{A. Velasquez was with University of Colorado Boulder. E-mail: \texttt{alvaro.velasquez@colorado.edu}}
\thanks{S. Khan was with the Air Force Research Laboratory. E-mail: \texttt{simon.khan@us.af.mil}}
}

\maketitle

\begin{abstract}  % MAX LENGTH 250 WORDS
State of the art reinforcement learning methods sometimes encounter unsafe situations. Identifying when these situations occur is of interest both for post-hoc analysis and during deployment, where it might be advantageous to call out to a human overseer for help.  Efforts to gauge the {\it criticality} of different points in time have been developed, but their accuracy is not well established due to a lack of ground truth, and they are not designed to be easily interpretable by end users.  Therefore, we seek to define a criticality framework with both a quantifiable ground truth and a clear significance to users.  We introduce {\it true criticality} as the expected drop in reward when an agent deviates from its policy for $n$ consecutive random actions. We also introduce the concept of {\it proxy criticality}, a low-overhead metric that has a statistically monotonic relationship to true criticality. {\it Safety margins} make these interpretable, when defined as the number of random actions for which performance loss will not exceed some tolerance with high confidence.  We demonstrate this approach in several environment-agent combinations; for an A3C agent in an Atari Beamrider environment, the lowest 5\% of safety margins contain 47\% of agent losses; i.e., supervising only 5\% of decisions could potentially prevent roughly half of an agent's errors.  This criticality framework measures the potential impacts of bad decisions, even before those decisions are made, allowing for more effective debugging and oversight of autonomous agents.
\end{abstract}

\begin{IEEEImpStatement}
To prevent autonomous agents from making catastrophic mistakes, they often require human oversight. In critical environments, this oversight has needed to be constant, as there have been no means of effectively identifying when a catastrophic mistake could occur. Previous methods attempted to identify these situations without any accuracy guarantees and without a meaningful quantification of risk. The methods introduced in this paper result in safety margins, or probabilistic bounds on the number of consecutive erroneous actions that would cause the agent's performance to drop by more than some user-specified amount. In an Atari Beamrider environment, 47\% of agent losses were in the lowest 5\% of safety margins. In other words, rather than providing constant oversight, supervising just 5\% of agent decisions could potentially prevent almost half of agent losses. This technology could assist in any number of autonomous applications, including, e.g., freight transport, drone operation, and automated manufacturing.
\end{IEEEImpStatement}


\section{Introduction}\label{sec:intro}

\IEEEPARstart{I}{n} reinforcement learning (RL), a policy is learned that dictates which immediate action should be executed for a given state, in order to maximize some long-term reward from an agent's environment. In some states, the action choice has no effect on the agent's expected reward: if an agent plays Pong, and the ball is far away, then it might not matter whether the paddle is moved up or down, as plenty of time exists to correct any instantaneously ``incorrect'' action and re-center the ball on the paddle \cite{lin2017}.  At other times, the choice of action is critical: if the ball is very close to the paddle, then a wrong move can result in the loss of a point. Quantifying this {\it criticality} at each moment in time can provide utility in a number of ways. 

One use case for such a {\it criticality metric} applies to a human analyst debugging an autonomous agent that failed its task -- e.g., losing a game of Pong or taking a series of detours that resulted in a 20 hour delay of a shipment. The {\it episode} describing an agent's task performance, i.e., the history of an agent's observations and actions, might be quite long, into the thousands or even millions of time steps. A well-defined and accurate criticality measurement could be used to draw attention to the relatively few decisions that had the greatest impact on task performance, greatly reducing the analyst's cognitive burden. Additional explanatory techniques could then be further used to solidify the analyst's understanding of that moment in time \cite{ribeiro2016,woods2019,huber2023}.

Another use case involves a human team actively managing one or more autonomous platforms, such as drones or self-driving trucks. In these critical applications, where the potential cost of failure is catastrophic, the oversight team needs to constantly survey the platforms to ensure optimal outcomes \cite{quan2019}. The introduction of real-time criticality measurements allows this team to better focus their efforts, paying attention when a wrong action will significantly affect episode outcomes (high criticality) and managing other tasks when a wrong action is inconsequential (low criticality). With this reduced cognitive burden for a single platform, one operating team could effectively multiplex their attention over multiple autonomous platforms. 

Over the last several years, various metrics have been proposed for estimating the criticality of actions executed at specific time steps \cite{lin2017,huang2018,spielberg2018,sun2020,guo2021,kumar2021,xu2021,spielberg2022}. For example, in \cite{lin2017}, some notion of criticality is estimated by looking at the probability distribution (or Q values) over possible action choices, and taking the difference between the probability of the best and worst actions. Their hypothesis was that if the difference in probabilities is greater, then the choice of action is more critical. Unfortunately, this metric, and other such metrics, suffer from two limitations.  First, the metrics can be erroneous: for example, they may output a low value for a situation that is critical (a false negative error), or a high value for a non-critical situation (a false positive); furthermore, no means are provided to understand how often these errors occur. Second, even when the output is correct, these metrics can be difficult to interpret in the context of any particular environment. In the example metric from \cite{lin2017}, the difference in probabilities does not directly indicate the severity of consequences that can be expected if some number of mistakes is made within a given application. 

In this paper, we seek to overcome these limitations. First, we establish a formal definition of {\it true criticality}:
\begin{align}
    &\text{true criticality} \nonumber\\ 
    &\quad := c(t, n; \pi) = \mathbb{E}_{a \sim \pi}\left[R_{\gamma}\right] - \mathbb{E}_{a \sim \pi'(t, n)}\left[R_{\gamma}\right], \label{eq:intro:criticality}
\end{align} 
fully defined in \cref{sec:methods:definition}. This equation expresses the expected reduction in total discounted reward that would result if the agent's chosen action (from its policy $\pi$) is randomly replaced for $n$ steps, beginning at time $t$; i.e., the agent's behavior is perturbed at time steps $t, t+1, ..., t+n-1$. This formulation strongly benefits from being in the cumulative reward space defined by the agent's environment; that is, the criticality units are automatically application-specific, and directly tied to whatever measurable outcomes exist, making them easy for end users to interpret.  A somewhat similar definition is informally mentioned in \cite{huang2018}, though in that study, a human user determines whether or not a state is critical -- a definition that both does not necessarily end up with units indicative of an application's outcomes, and cannot be automated; the definition also considers perturbations only at a single time step ($n = 1$). In contrast, the definition proposed here can be automatically and tractably estimated as a result of the expectation operator in \cref{eq:intro:criticality}. We provide what is, to our knowledge, the first fully-automated method for tractably approximating true criticality with a high degree of accuracy. However, the estimation cannot typically be performed in real-time, since it involves running repeated simulations of the future.

Next, we show how existing metrics (including \cite{lin2017}) can be viewed as {\em proxy criticality} metrics:
\begin{align}
    \text{proxy criticality} := p(t, n; \pi) = f(o_t, o_{t-1}, ..., n; \pi), \label{eq:intro:proxy},
\end{align}
fully defined in \cref{sec:methods:proxy}.
While these proxy criticality metrics have indeterminate accuracy due to their lacking of a well-grounded quantitative definition, they can be very efficiently computed, providing useful measurements in real-time.

We present methods for analyzing the statistical relationship between the true criticality and proxy criticality metrics. These methods combine the advantages of each (accuracy and speed, respectively) into a framework that maps a quick-to-compute proxy metric value at some time $t$ (from \cref{eq:intro:criticality}) to a distribution over true criticality values, for different numbers of random steps ($n$ from \cref{eq:intro:criticality}, roughly analogous to the number of potential mistakes).  Then, the statistical binding between true criticality and proxy criticality metrics is exploited to compute {\em safety margins}, defined as the number of potentially incorrect actions that can be tolerated, with high confidence, before the expected impact on an agent's outcomes (i.e., true criticality) surpasses some user-defined tolerance $\zeta$. Our hypothesis is that even with a proxy criticality metric that has limited accuracy, we would be able to generate meaningful safety margins with probabilistic guarantees. 

To test this hypothesis, we analyze and demonstrate the above techniques within two classic Atari games: Pong and Beamrider, with models trained using the APE-X \cite{horgan2018} and A3C \cite{mnih2016} algorithms, and with a criticality metric adopted from \cite{lin2017}.  We show that we can establish a clear relationship between the metric and safety margins, despite significant limitations to how accurately this metric can predict true criticality. Importantly, the framework for relating other proxy criticality metrics to the proposed true criticality metric provides a basis for comparing different proxy metrics in the future. We empirically validate our approach by showing that the safety margins are typically accurate, and that as expected, they decrease significantly as the agent approaches the point in time at which it loses the game -- in Beamrider with an A3C agent, for example, \SI{47}{\%} of agent losses happen in the \SI{5}{\%} smallest safety margins.  Our experimental results support the hypothesis given earlier, and suggest that our approach can potentially benefit both of our identified use cases: drawing user attention to crucial moments in an episode post facto, and supporting real-time indicators of when intervention from human teams overseeing autonomous platforms are likely to be most appropriate.

We note that the safety margin generation approach was summarized in a two page publication \cite{grushin2023}, but is presented in much greater technical detail here, with previously-unpublished analysis and results.

\section{Background}\label{sec:background}

As mentioned earlier, a number of metrics related to criticality have been developed.  Some metrics have definitions that vary by domain, requiring additional design work and testing for new domains; for example, in \cite{spielberg2022}, a custom metric is given for the Pong environment that explicitly takes into account the proximity of the ball to the agent's paddle.  Other metrics are domain-independent, including the aforementioned criticality metric of \cite{lin2017}, which can be extended to, e.g., the entropy of probability distribution output by the policy or the difference between the maximum Q value and the average Q value \cite{huang2018}. In \cite{guo2021}, a more complex approach is given, where a separate model is trained on a large number of episodes to predict the total reward that an agent will gain for a given episode; the trained model is then analyzed to determine the extent to which different actions have influenced the prediction. In this approach, called EDGE, the degree of influence of an action upon total reward is treated as an estimate of its criticality; EDGE omits any statistical guarantees on the accuracy of this proxy metric.

Concurrently with the development of criticality metrics, researchers have attempted to establish procedures for evaluating the metrics' performance \cite{lin2017,huang2018,sun2020,guo2021,kumar2021}.  In one approach, the evaluation is performed by a human user, who determines whether a particular time step is critical; the criticality metric's output is compared against the determination that was made by the user, and the degree of accuracy (agreement) allows the user to determine the extent to which the agent can be trusted \cite{huang2018}.  Another approach comes from the adversarial attack context, wherein an attacker might select all time steps where the criticality (as estimated by the metric) is particularly high, to fool the agent into executing suboptimal actions at those time steps. Then, the total reward collected by the agent throughout the episode is observed; for a given number of time steps selected, a lower reward indicates a more accurate metric \cite{lin2017,sun2020,kumar2021}.  For the purposes of explainable learning, an evaluation procedure is proposed in \cite{guo2021}, where it is used to compare the EDGE metric against several other criticality metrics (including, in limited scenarios, the criticality metric of \cite{lin2017}, which EDGE outperforms).  Rather that adversarially perturbing the observation at each selected time step, the procedure simply replaces the action that is prescribed by the agent's policy with a randomly-chosen action -- a much more computationally efficient approach, which nonetheless uses similar evaluation criteria of maximum impact for a small number of changed actions.

These prior approaches focus on penalizing false positives in $p(t, n; \pi)$, wherein a high metric value does not correspond to a significant impact on the episode's outcome. They tend to ignore false negatives, where the metric value is low but the impact on the episode is significant. This issue stems from these evaluation metrics being defined independent of any ground truth or gold standard; i.e., true criticality. 

\section{Methodology}\label{sec:methods}

Our methodology is divided into three steps: first, defining the true criticality in \cref{sec:methods:true}; second, defining the family of proxy criticalities in \cref{sec:methods:proxy}; and finally, defining safety margins as a result of analyzing the statistical relationship between these quantities in \cref{sec:methods:safetymargins}.  Experimental details are given in \cref{sec:app:setup}.

\subsection{True Criticality}\label{sec:methods:true}

In the subsections below, we provide a formal definition of true criticality (introduced in \cref{eq:intro:criticality}), describe a procedure for its tractable approximation, and establish an approach for controlling the degree of approximation error.

\subsubsection{Definition}\label{sec:methods:definition}

We formally define true criticality as follows:

Consider a policy $\pi$ that was learned by some agent, and that outputs an action $a$ that the agent considers to be best, given an observation $o$.  Then, consider a {\it perturbed policy} $\pi'(t, n)$, which is identical to $\pi$, except that at each of time steps $t, t+1, ..., t+n-1$, $\pi'(t, n)$ outputs an action chosen uniformly at random (including the choice that $\pi$ would have made).  Let $\mathbb{E}_{a \sim \pi}\left[R_{\gamma}\right]$ be the expected sum $R_{\gamma} = \sum_{k=t}^{\infty} \gamma^{k-t} r_{k}$ of individual $\gamma$-discounted rewards $r_{k}$ obtained by the agent if it executes the actions prescribed by its policy $\pi$, beginning at time step $t$, and proceeding until the end of the episode (note that the quantity $R_{\gamma}$ depends upon the subscript of the expectation operator $\mathbb{E}$).  
Then the {\it true criticality} $c(t, n; \pi) = \mathbb{E}_{a \sim \pi}\left[R_{\gamma}\right] - \mathbb{E}_{a \sim \pi'(t, n)}\left[R_{\gamma}\right]$ is the difference between $\mathbb{E}_{a \sim \pi}\left[R_{\gamma}\right]$ and the expected total discounted reward $\mathbb{E}_{a \sim \pi'(t, n)}\left[R_{\gamma}\right]$ obtained by the agent if it executes the actions prescribed by the perturbed policy $\pi'(t, n)$.

In this definition, rewards before $t$ do not need to be considered, since their sampled values are identical for both $\pi$ and $\pi'(t, n)$.  In \cref{sec:app:alternativedefinitions}, we compare this definition with several other possible definitions for true criticality, and argue that it is more appropriate or practical than the considered alternatives, though we do consider possible future variations in \cref{sec:future}.

\subsubsection{Approximation}\label{sec:methods:approximation}

\begin{algorithm}[b!]
    \SetKwInOut{Input}{input}
    \Input{A time step, $t$; a number of actions to perturb, $n$; a number of trials for expectation, $N$}
    Run an episode for $t$ time steps, starting with the initial observation $o_{0}$, to observation $o_{t}$, following policy $\pi$. \\
    From time step $t$, compute the unperturbed reward: run the episode for $h$ time steps, following policy $\pi$, and measure the total discounted reward $\sum_{k=t}^{t+h-1} \gamma^{k-t} r_{k}$. \label{sec:methods:approximation:unperturbedreward} \\
    Compute the estimated expected perturbed reward $\mathbb{E}^*_{a \sim \pi'(t, n)}\left[R_{\gamma, h}\right]$ from perturbed reward values collected over $N$ independent trials, with $\sum_{k=t}^{t+h-1} \gamma^{k-t} r_{k}$ computed in each trial while following policy $\pi'(t, n)$. \label{sec:methods:approximation:meanperturbedreward} \\
    Compute the mean reward reduction, or approximate true criticality: $c^*(t, n; \pi) = \mathbb{E}^*\left[\Delta R_{\gamma, h}\right]$ by subtracting the result of \cref{sec:methods:approximation:meanperturbedreward} from the result of \cref{sec:methods:approximation:unperturbedreward}.
\caption{True Criticality Approximation} \label{alg:approximation}
\end{algorithm}

We now describe an algorithm for approximating true criticality, according to the definition.  In many environments, it may be prohibitively expensive to compute the true criticality values $c(t, n; \pi)$ exactly: if an agent can execute $|A|$ actions at every time step, then there are $|A|^{n}$ possible perturbed action sequences, and the expected perturbed reward $\mathbb{E}_{a \sim \pi'(t, n)}\left[R_{\gamma}\right]$ must be computed over all of these sequences.  Furthermore, if the environment is stochastic, the same sequence of actions may lead to different rewards, requiring multiple samples from each sequence.

To avoid high computational costs, we approximate the expected value $c(t, n; \pi) = \mathbb{E}_{a \sim \pi}\left[R_{\gamma}\right] - \mathbb{E}_{a \sim \pi'(t, n)}\left[R_{\gamma}\right]$ as a sample mean $c^*(t, n; \pi) = \mathbb{E}^*\left[\Delta R_{\gamma}\right] = \mathbb{E}^*_{a \sim \pi}\left[R_{\gamma}\right] - \mathbb{E}^*_{a \sim \pi'(t, n)}\left[R_{\gamma}\right]$ that is computed by repeatedly running an episode with different random action choices made at time steps $t, t+1, ..., t+n-1$ over some number $N$ of trials, and for each trial, measuring the difference in total discounted reward between the perturbed trial and the original episode.  (Note that here, and elsewhere in the paper, the $*$ symbol is used to denote a statistic that is approximated from a sample, rather than an ideal, ``population''-based statistic.)  As an additional optimization, we introduce a finite horizon $h$ and only measure discounted rewards for $h$ steps; this is applied both to the original, unperturbed episode, and the perturbed trials, with the difference denoted by $\mathbb{E}^*\left[\Delta R_{\gamma, h}\right]$ (note the additional $h$ subscript). The error introduced by this optimization is bounded due to the discount factor, $\gamma$, as discussed in \cref{sec:methods:analysis}.

The criticality approximation procedure is listed as \cref{alg:approximation}. Two key assumptions are made, but adjustments are recommended for situations where one of these assumptions does not hold (though in the case where neither assumption holds, the true criticality cannot be estimated with the given method):

\begin{enumerate}[I)]
    \item The environment is deterministic: given the same initial state/observation $o_{0}$, the same sequence of actions will yield the same sequence of states/observations.  If this assumption does not hold, then the computation of total discounted unperturbed reward in \cref{sec:methods:approximation:unperturbedreward} should be performed over $N$ trials, as is done for the perturbed reward in \cref{sec:methods:approximation:meanperturbedreward}.
    \item The environment's state/observation at time $t$, $o_t$, can be saved and loaded/replayed multiple times, in order to gather $N$ perturbed reward measurements for $\mathbb{E}^*\left[\Delta R_{\gamma, h}\right]$.  If this second assumption does not hold, but the initial simulator state can be saved and the environment is deterministic, then in \cref{sec:methods:approximation:meanperturbedreward}, the episode should be run from observation $o_{0}$, rather than observation $o_{t}$, replaying all actions up to time $t$. This is less efficient, but achieves the same effect.
\end{enumerate}

\subsubsection{Error Analysis and Control} \label{sec:methods:analysis}

We identify two types of {\it approximation error} in the true criticality calculation from \cref{alg:approximation}: {\it horizon error}, which arises as a result of not running an episode to completion before computing the total discounted reward accumulated, and {\it sampling error}, as a result of computing the mean reward reduction over a fixed number of trials.

\paragraph{Horizon error} We define horizon error in as the relative magnitude of the difference between the mean reward reduction computed with or without a horizon. Then:
\begin{align}
    \epsilon_{horizon} &= \left( \mathbb{E}^*\left[\Delta R_{\gamma, \infty}\right] - \mathbb{E}^*\left[\Delta R_{\gamma, h}\right] \right) / \mathbb{E}^*\left[\Delta R_{\gamma, \infty}\right] \\
        &= \frac{\sum_{k=t+h}^\infty \gamma^{k - t} \Delta r_{k}}{\sum_{k=t}^\infty \gamma^{k - t} \Delta r_{k}}.
\end{align}

The individual reward changes $\Delta r_{k}$ (at different time steps $k$) can be viewed as samples of a random variable.  Under the assumption that the environment's rewards have a stationary distribution (that is independent of the specific time step $k$), the $\Delta r_{k}$ terms can be replaced with their expected value $\mathbb{E}\left[\Delta r\right]$, computing the expected horizon error as:
\begin{align}
    \mathbb{E}\left[\epsilon_{horizon}\right] &= \frac{\mathbb{E}\left[\Delta r\right] \sum_{k=t+h}^\infty \gamma^{k - t}}{\mathbb{E}\left[\Delta r\right] \sum_{k=t}^\infty \gamma^{k - t}} = \gamma^h.
\end{align}

Conveniently, this expresses the fact that the number of time steps needed to achieve a desired level of error is a constant dependent only on $\gamma$ (irrespective of environment or application). The assumption that rewards follow a stationary distribution is a limitation of this approach; however, in practice, this can be dealt with by choosing a horizon that sufficiently limits $\epsilon_{horizon}$, in turn causing any significant rewards outside of the horizon to be scaled by an arbitrarily low $\gamma^h$.
Given some desired error threshold $\hat{\epsilon}_{horizon}$, we can rearrange and discretize the above to obtain a method for selecting $h$:
\begin{align} \label{eqn:horizon}
h &= \lceil \log_{\gamma} \hat{\epsilon}_{horizon} \rceil.
\end{align}

\paragraph{Sampling error} 
We define sampling error, $\epsilon_{sampling}$, as the difference between the measured mean reward reduction $\mathbb{E}^*\left[\Delta R_{\gamma, h}\right]$ and the true expected reward reduction, $\mathbb{E}\left[\Delta R_{\gamma, h}\right]$. Under the assumption that the distribution of sample reward reductions is normal, we can follow the procedure from \cite{driels2004} to bound this error. With a confidence of $\alpha$, the following holds, given the appropriate constant $t_\alpha$ for a Student's t-distribution:
\begin{align}
\lvert \epsilon_{sampling} \rvert &\leq \frac{t_{\alpha} \mathrm{stdev}^*\left[\Delta R_{\gamma, h}\right]}{\sqrt{N}},
\end{align}\label{eqn:sampling}
where $\mathrm{stdev}^*\left[\Delta R_{\gamma, h}\right]$ is the sample standard deviation (with Bessel's correction) of the observed reward reductions across $N$ trials.  As each $t$ within an episode will have a widely varying $\mathrm{stdev}^*\left[\Delta R_{\gamma, h}\right]$, the number of trials $N$ needed for each true criticality estimate may also vary significantly.
Therefore, to choose a number $N$ that approximates the true criticality within some target $\hat{\epsilon}_{sampling}$ range, the right hand side of the above inequality is computed after executing each trial, and \cref{sec:methods:approximation:meanperturbedreward} of \cref{alg:approximation} is terminated only once the right hand side is less than or equal to the target $\hat{\epsilon}_{sampling}$.  For sufficiently accurate estimates of $\mathrm{stdev}^*\left[\Delta R_{\gamma, h}\right]$, some minimum number $N_{min}$ of trials is always performed. 

Notably, the units of $\hat{\epsilon}_{sampling}$ are specific to the rewards from a given environment. That is, they are defined in absolute terms, rather than as a unitless, relative quantity (as was the case for $\epsilon_{horizon}$).

\subsection{Proxy Criticality}\label{sec:methods:proxy}

Even with the optimizations and assumptions discussed in \cref{sec:methods:approximation}, approximating the true criticality is prohibitively expensive to do in real-time. However, there are many metrics that {\em can} be computed in real-time and may be related to true criticality. We propose that an efficient way to estimate true criticality in real-time is to find a proxy metric that has an approximately monotonic relationship with the true criticality. Any of these proxy metrics may be formulated as $p(t, n; \pi)$ (see \cref{eq:intro:proxy}), and are constrained to only look at the history of agent observations, $o_t, o_{t-1}, ...$, and the number of perturbed actions, $n$. To our knowledge, existing proxy criticality metrics have not considered $n$ as a parameter, and most rely only on the present observation $o_t$, rather than past observations.  We note that future work on, e.g., world model-derived policies could easily integrate this information \cite{ha2018,schrittwieser2020}.  (For such sampling-based approaches, we would denote the proxy metric by $p^*(t, n; \pi)$, following the convention used in this paper.)

For the purposes of this study, we used variations of the criticality metric of \cite{lin2017}, which is not sampling-based.  Specifically, when action probabilities are available (as in the case of A3C), we defined $p(t, n; \pi)$ as the difference between the largest and smallest log probabilities (rather than raw probabilities) across all possible action choices at time $t$.  When action probabilities are not available (as in the case of APE-X), we instead computed the difference between the largest Q value and the smallest Q value.

\subsection{Safety Margins}\label{sec:methods:safety}

We now combine the true and proxy criticalities to obtain safety margins. As mentioned in \cref{sec:intro}, safety margins describe the number of potentially incorrect actions that can be tolerated before the expected impact on an agent's outcomes surpasses some user-defined tolerance $\zeta$. We first give a formal definition of safety margins, then describe how proxy and true criticality data can be collected and used for safety margin approximation, and finally, discuss sources of approximation error for safety margins.

\subsubsection{Definition}\label{def:safetymargins}

We define safety margins, discussed in \cref{sec:intro}, as follows:

Let the {\it $\zeta$-tolerance safety margin} $s(t, \zeta; \pi) = \argmax_{n}{ \forall_{n' \in \{0, 1, 2, ..., n \}} \left[ c(t, n'; \pi) \leq \zeta \right] }$ be the maximum number of time steps $n$ (beginning with time step $t$) for which actions can be perturbed without reducing the expected total discounted reward by more than $\zeta$.

Then, we define the {\it estimated $\zeta$-tolerance safety margin} $s^*(t, \zeta; \pi) = \argmax_{n}{ \forall_{n' \in \{0, 1, 2, ..., n \}} p(t, n'; \pi) \implies \left[ c^*(t, n'; \pi) \leq \zeta \right]}$, where the $\implies$ symbol denotes that given some current proxy metric value $p(t, n'; \pi)$, the desired constraint $c^*(t, n'; \pi) \leq \zeta$ will be respected with high probability.

Following are a few notes on these definitions. We can approximately view the maximum number of random action perturbations $n$ as the number of mistakes that an agent can afford to make, with the understanding that with $|A|$ choices, there is a $\frac{1}{|A|}$ probability that the random action choice will be the same as the action choice output by the policy $\pi$, and will therefore not be a mistake.  It is also worth noting that by the definition of true criticality given in \ref{sec:methods:definition}, $c(t, 0; \pi) = 0$ (if no actions are perturbed, then the expected reduction in reward is $0$).  Thus, if perturbing any number of actions (even just one, at time $t$) will result in an expected reward reduction that is greater than $\zeta$, then the safety margin will be $0$.

Ideally, $c(t, n; \pi)$ would be monotonically non-decreasing with respect to $n$. However, as $\pi$ is not necessarily a perfect policy, and since the approximation $c^*(t, n; \pi)$ of $c(t, n; \pi)$ from \cref{sec:methods:approximation} is subject to probabilistic noise (albeit noise that can be controlled as described by \cref{eqn:sampling}), this is not always the case. This motivates the $\forall$ clause in the definition, forcing estimates at {\em} any value $n \leq s(t, \zeta; \pi)$ to be below the user specified tolerance, $\zeta$.

\subsubsection{Data Collection}\label{sec:methods:collection}
\begin{algorithm}[b!]
    \SetKwInOut{Input}{input}
    \Input{A set of perturbation sizes, $S$; a number of data tuples to collect, $M$}
    \For{each episode $m \in [1, 2, ..., M]$}{
        Run the episode to completion, starting with the initial observation $o_{0}$, following policy $\pi$, and obtain the proxy criticality value $p_{m}(t)$ for each time step $t$. \\
        \eIf{$m = 1$ or {\it uniform sampling} is not desired}{%
            Select $t_m$ randomly and uniformly.
        }{%
            Select $t_m$ that maximizes the minimum absolute difference $\min_{\mu}| p_{m}(t_{m}) - p_{\mu}(t_{\mu}) |$ between $p_{m}(t_{m})$ and the closest proxy criticality value $p_{\mu}(t_{\mu})$ for any time step $t_{\mu}$ that was selected for some previous episode $\mu < m$, with ties broken randomly and uniformly.
        }
        \For{$n \in S$}{
            Compute the estimated true criticality $c_{m}^{*}(t_{m}, n; \pi)$ using \cref{alg:approximation}.
        }
    }
    \caption{Safety Margin Data Collection} \label{alg:evaluation}
\end{algorithm}
As mentioned earlier, safety margins are approximated via the relationship between proxy and true criticality values.  In order to capture this relationship, we run many episodes, collecting a data tuples of the following form from each:
\begin{align*}
    ( \qquad & p_{m}(t_{m}, n_1; \pi), \quad ..., \quad p_{m}(t_{m}, n_{|S|}; \pi), \\
        & c_{m}^{*}(t_{m}, n_{1}; \pi), \quad ..., \quad c_{m}^{*}(t_m, n_{|S|}; \pi)  \qquad ).
\end{align*}

Each tuple consists of a proxy value for some episode $m$, at some specific time step $t_{m}$, along with estimated true criticality values at $t_{m}$, for different values $n$; the set of $n$ values evaluated is denoted by $S = \{n_{1}, n_{2}, ..., n_{|S|}\}$. 
We add the subscript $m$ to a variable to denote that its value is computed specifically for episode $m$ (e.g., writing $p_{m}(t_{m}, n_1; \pi)$ instead of $p(t, n_1; \pi)$).  Additionally, in cases where $p(t, n; \pi)$ does not depend on $n$, all proxy criticalities are equal, and the tuple can elide all but one, written as $p_{m}(t_{m}; \pi)$, or even $p_{m}(t_{m})$, for brevity.

The procedure for collecting data tuples is listed as \cref{alg:evaluation}. For each episode $m$, we select a time step $t_{m}$ at which to apply the criticality metric $p_{m}(t_{m})$ and to compute the estimated true criticality $c_{m}^{*}(t_{m}, n; \pi)$. The time step selection algorithm attempts to provide a relatively uniform sampling of proxy criticality values, which can be non-trivial: in the environments that we consider, we found that for most time steps, proxy criticality is very low (see \cref{fig:results:metrichistogram:beamrider}).  To mitigate this, the algorithm keeps track of the criticality values $p_{\mu}(t_{\mu})$ for time steps $t_{\mu}$ that have been selected thus far, for earlier episodes $\mu < m$, and selects the time step $t_{m}$ for which the criticality value is as far as possible from any previously-selected criticality value; we illustrate this approach on a simple example in \cref{sec:app:setup} (see \cref{fig:methods:evaluation:sampling}).  A potential drawback of only using this approach is that it would undersample the many time steps $t_{m}$ that have a very low criticality value $p_{m}(t_{m})$, and may thus capture few false negative errors, where the $p_{m}(t_{m})$ is low, but the true criticality $c_{m}^{*}(t_{m}, n; \pi)$ is high.  To address this, we select $t_{m}$ uniformly both across the natural data distribution and across the space of proxy metric values, combining these samplings into one data pool for the computation of the estimated safety margins.

\subsubsection{Calculation}\label{sec:methods:safetymargins}
Once tuples containing proxy and true criticality values have been collected, the relationship between them can be analyzed to allow for the real-time estimation of safety margins for different values of $\zeta$.

For each value $n \in S$, we capture the relationship between values of $p(\cdot)$ and $c^*(\cdot)$ via a 2d {\em kernel density estimate} (KDE) plot, marginalizing out $t$ via matched tuples collected in \cref{sec:methods:collection}, and thus replacing the $t$ parameter with $\cdot$ (we also omit the $n$ and $\pi$ parameters for brevity).  Proxy criticality is placed on the horizontal axis, and true criticality on the vertical axis, with the axes discretized into a large number of small ``bins''`.  For each bin, the sum of true criticality density values are normalized such that their sum is $1$. Now, given a specific proxy criticality value $p(\cdot)$, the corresponding bin has an estimated probability distribution across different estimated true criticality values, $c^*(\cdot)$.

Within the probability distribution for a given proxy criticality metric value and for some value $n \in S$, we compute some high (e.g., $95$th) percentile $b_{\beta}^{*}\left( p(\cdot), n \right)$ for true criticality; i.e., $b_{\beta}^{*}\left( p(\cdot), n \right)$ is chosen such that $p(\cdot) \implies c^*(\cdot) \leq b_{\beta}^{*}\left( p(\cdot), n \right)$ with probability $\beta$. We can view the percentile $b_{\beta}^{*}\left( p(\cdot), n \right)$ as a probabilistic upper {\it bound} on the true criticality associated with a specific proxy criticality $p(\cdot)$.  

Finally, we approximate the safety margin $s^*(p(\cdot), \zeta; \pi, S)$ as follows, now expressing it as a function of $p(\cdot)$, rather than $t$ (note also that the $S \bigcup \{0\}$ statement allows for safety margins of $0$, even when $0 \notin S$):
\begin{align*}
    &s^*(p(\cdot), \zeta; \pi, S) \\&\quad%\gtrapprox_{\beta} s(p(\cdot), \zeta; \pi, S) \\
        = \argmax_{n}{ \forall_{\{n' \in S \bigcup \{0\} : n' \leq n\}}} [ b_{\beta}^{*}\left(p(\cdot), n'\right) \leq \zeta ].
\end{align*}

This estimated safety margin $s^*(p(t, n; \pi), \zeta; \pi, S)$ can be viewed as a probabilistic lower bound on the true safety margin $s(t, \zeta; \pi)$ at some time $t$, and is expected to hold with probability approximately $\beta$. The choice of both $\beta$ and $\zeta$ depend upon the needs of the specific application, including the risk tolerance of a given user.

\subsubsection{Error Analysis and Control} \label{analysis:safetymargins}

In the following, we discuss sources of error that exist in the approximation of safety margins, and how these errors can be mitigated.

\paragraph{Criticality error}
We note that the errors in true criticality (\cref{sec:methods:analysis}) propagate to the approximation of safety margins, but they can be accounted for by selecting the appropriate bandwidth on the vertical axis of the 2d KDE plots, since bandwidth captures uncertainty in the underlying data.  In particular, assuming that horizon error is low enough to be ignored, the vertical bandwidth (standard deviation) of a 2d Gaussian kernel can be set to the standard deviation $\hat{\epsilon}_{sampling} / t_{\alpha}$ of the sample mean $\mathbb{E}^*\left[\Delta R_{\gamma, \infty}\right]$ (or to a larger value), assuming that the sample size $N$ is large enough that the sample mean follows an approximately normal distribution.

\paragraph{Percentile error}
Safety margin approximation has its own source of sampling error, stemming from the fact that the number of episodes processed by \cref{alg:evaluation} (and thus the number of data tuples generated) $M$ is finite.  This sampling error affects the estimated percentiles $b_{\beta}^{*}\left(p(\cdot), n\right)$, from which safety margins are directly derived.  The true, population percentile, which we denote via $b_{\beta}\left(p(\cdot), n\right)$, is unknown, since an infinite number of tuples would be needed in order to compute it.  Now, define $\beta^{*}$ such that $b_{\beta^{*}}\left( p(\cdot), n \right) = b_{\beta}^{*}\left( p(\cdot), n \right)$; in other words, $\beta^{*}$ is chosen such that the $\beta^*$ percentile in the true (population) distribution is equal to the $\beta$ percentile in the estimated (sample) distribution.  Then, we define {\it percentile error} $\epsilon_{percentile}$ as follows:
\begin{align}
\epsilon_{percentile} &= \beta - \beta^{*}.
\end{align}

Because the probability distribution is estimated via a 2d KDE plot, it can be difficult to generate an accurate bound on percentile error.  However, as we show in \cref{sec:app:percentileerror}, under mild assumptions, $\epsilon_{percentile}$ decreases with the square root of $M$; thus, percentile error can be effectively controlled by generating a sufficient number of data tuples.  In \cref{sec:results}, we empirically evaluate the quality of the percentiles, and show that the errors are reasonably low.

\paragraph{Gap error}
Safety margin estimates can be affected by gaps within the set $S$, which specifies the values $n$ for which true criticality is computed.  This {\it gap error} can be quantified as follows:
\begin{align}
\epsilon_{gap} &= s^*(p(\cdot), \zeta; \pi, S) - s^*(p(\cdot), \zeta; \pi, \mathbb{N}).
\end{align}

Here, $\mathbb{N}$ is the set of natural numbers.  To illustrate the error, suppose that $S = \{1, 2, 4 \}$, and that $c(t, n; \pi) \leq \zeta$ for $n = 1$, $n = 2$, and $n = 3$, but not for $n = 4$.  Then, since $3 \notin S$, the computed $\zeta$-tolerance safety margin will be $2$, even though it is actually $3$; in other words, the safety margin is more conservative than it could be.  The opposite situation, which is more serious, can occur if true criticality $c(t, n; \pi)$ is not a monotonically-increasing function of $n$.  Consider the aforementioned example, but suppose that now, $c(t, 4; \pi) \leq \zeta$ but $c(t, 3; \pi) > \zeta$.  Then, even with a perfect proxy criticality metric (which predicts the true criticality exactly), our approximation will result in an estimated safety margin of $s^*(p(\cdot), \zeta; \pi, S) = 4$ for time step $t$, even though the true safety margin $s(t, \zeta; \pi)$ should be $2$, according to the definition in \cref{def:safetymargins} (note the $\forall$ statement in the definition).  To mitigate this error, more integers can be included in $S$, though this does increase the computational overhead for \cref{alg:evaluation}.

\section{Results/Analysis/Discussion}\label{sec:results}

The proposed methods were validated empirically in two classic Atari environments: Pong, a simple emulation of table tennis; and Beamrider, a fixed shooter game.  For each environment, two models were trained: one using the APE-X algorithm \cite{horgan2018}, based on the deep Q-network approach (DQN); and another model using the A3C algorithm \cite{mnih2016}.  In the following, we begin by presenting the computed safety margins, along with the underlying KDE plots and percentile curves.  For brevity, we include only the full KDE plot as generated from an A3C agent trained on Beamrider, and a safety margin plot for APE-X and A3C agents trained on Beamrider; other such figures (particularly, ones for Pong), may be found in \cref{sec:app:results}.

We then perform a statistical validation of the percentile curves, by generating them from a subset of the data tuples with $\beta = 0.95$, and then showing that for the remaining data tuples, the likelihood of true criticality exceeding the percentile is usually not far above (and often below) the desired value of $5\%$.  For Beamrider, we also show that safety margins decrease when the agent is about to lose the game.  Finally, we discuss our approach and results within the context of existing work.

\begin{figure*}[t]
    \centering
    \includegraphics[clip, width=\linewidth]
    {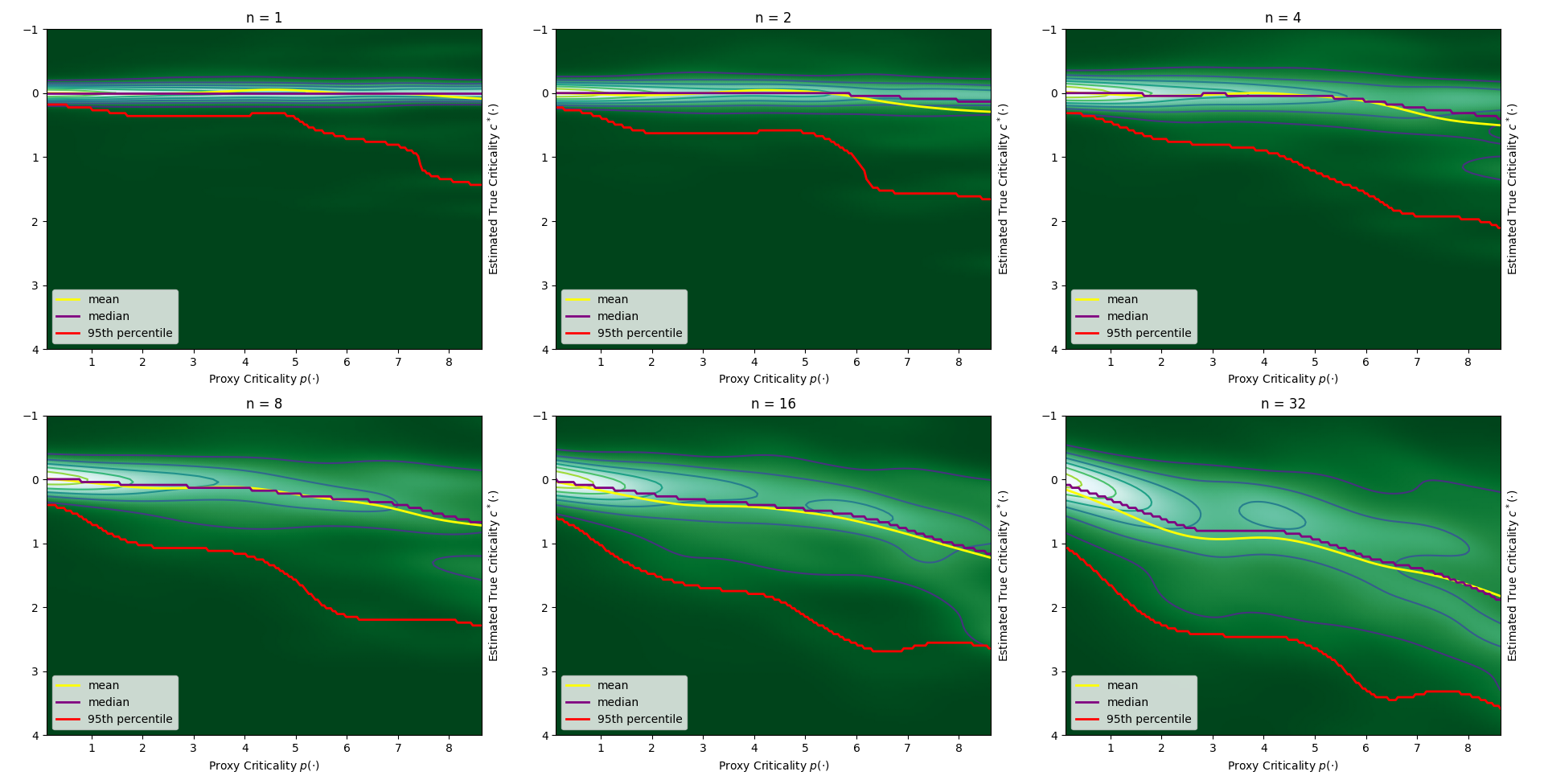}
    \caption{Normalized KDE plots capturing the relationship between proxy criticality (on the horizontal axis) and estimated true criticality (on the vertical axis), with contours and trendlines, for an A3C model trained on Beamrider.  A portion of this figure appeared in \cite{grushin2023}, with some differences in notation.}
    \label{fig:results:kde:a3c:beamrider}
\end{figure*}

\begin{figure}[t]
    \centering
    \includegraphics[clip, width=0.51\linewidth]{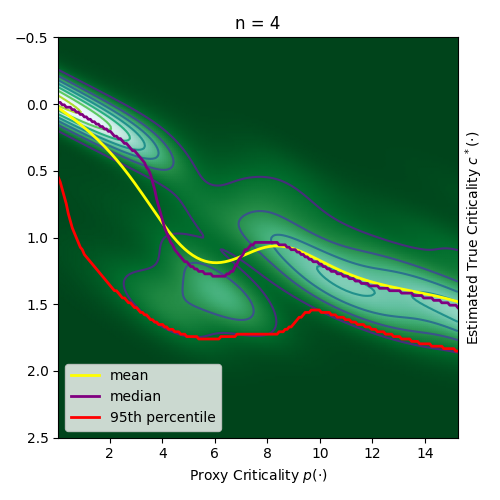}
    \caption{Normalized KDE plot excerpt for an A3C model trained on Pong; see \cref{sec:app:results} (\cref{fig:results:kde:a3c:pong}) for the full figure.}
    \label{fig:results:kde:a3c:pong:excerpt}
\end{figure}

\begin{figure*}[t]
    \centering
    \includegraphics[clip, width=\linewidth]
    {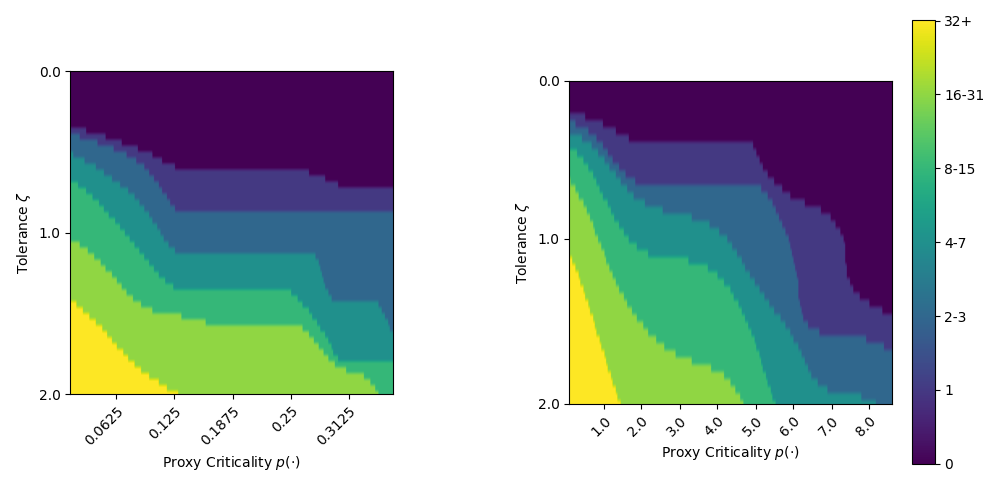}
    \caption{Heatmaps displaying safety margins for Beamrider in the proxy criticality metric vs. tolerance space, for APE-X (left) and A3C (right).  This figure was originally presented in \cite{grushin2023}, albeit with some differences in notation.}
\label{fig:results:safetymargins:beamrider}
\end{figure*}

\subsection{True and Proxy Criticalities}

The KDE plots for an A3C agent playing Beamrider are displayed in \cref{fig:results:kde:a3c:beamrider}, capturing the relationship between proxy criticality and estimated true criticality, for $S = \{1, 2, 4, 8, 16, 32\}$.  These (and other KDE plots) were computed by first generating $M = 1,000$ data tuples, but then removing the $5\%$ of the data tuples with highest proxy criticality values; this was done to ensure that, where safety margins are later generated, the probabilistic guarantees are statistically significant.  Mean, median and $95$th percentile curves $b_{\beta=0.95}^{*}\left( p(\cdot), n \right)$ are then computed (see \cref{sec:methods:safetymargins}).  The plots and curves demonstrate characteristics of the Beamrider environment -- for example, the $n=1$ subplot shows that individual random actions have a very limited effect on the agent's performance unless the proxy criticality is greater than $5$.  Or: any single mistake is unlikely to significantly affect the agent's performance in situations with low proxy values. On the other hand, at $n=16$, even proxy values as low as 1 have a noticeably higher true criticality, showing that $16$ random actions is usually enough to impact the agent's performance.

For an example of a KDE that is less smooth, \cref{fig:results:kde:a3c:pong:excerpt} shows the $n=4$ KDE for the A3C agent playing Pong. In that figure, the true criticality is much less smooth than in Beamrider, and the relationship between true and proxy criticality does not appear to be strictly monotonic. This likely stems from a few factors: (1) the impact of any action on Pong is limited to either +1 for winning a point, 0 for continuing the game, or -1 for losing a point, with true criticality displaying intermediate values based on a probability distribution over these options; and (2) the proxy metric comes from a neural network, and that neural network appears to underestimate the importance of situations corresponding to a proxy value between $4$ and $8$.
While the trained agent has a mean reward between $20$ and $21$ (with $21$ being the maximum possible reward), the agent's overall performance does not necessarily reflect the quality of its proxy metric.  When computing safety margins, we mitigate the effects of such flaws by forcing the percentile line to be monotonic, as we describe below.

\subsection{Safety Margins}

Next, the $95$th percentile curves are used to perform safety margin estimations $s^*(p(\cdot), \zeta; \pi, S)$. The margin estimations are based on lookup tables generated from the KDEs. The Beamrider agents' safety margin lookup tables are shown as heatmaps in \cref{fig:results:safetymargins:beamrider}.  These heatmaps are converted into safety margin estimates by finding the current proxy criticality value $p(t, n; \pi)$ for time $t$ on the horizontal axis, and a desired, application-specific tolerance $\zeta$ on the vertical axis. 
Very high values of $p(\cdot)$ may fall outside of the range of the heatmap's horizontal axis, as we filter out $5\%$ of the data tuples with the highest proxy criticality values before computing the kernel density plots and safety margin lookup tables.

In each heatmap, the boundaries between shaded regions are formed by the $95$th percentile curves (\cref{fig:results:kde:a3c:beamrider}), except that these curves are first adjusted to ensure a monotonic relationship with $p(\cdot)$; details are given in \cref{sec:app:setup}.  This prevents non-monotonicity (e.g., as observed in \cref{fig:results:kde:a3c:pong:excerpt}) from affecting the safety margin's guarantees. This monotonic adjustment only ever increases a percentile estimate, ensuring that there will always be at least a $95\%$ confidence that the impact on application performance, given some number of random steps, will not exceed $\zeta$.

Given the adjusted $95$th percentile curves, we use purple shading for any point on the heatmap that is bounded above by the $95$th percentile curve for $n = 1$; this region corresponds to a safety margin of $0$.  Then, for each additional value of $n$ (in increasing order), we assign a specific color to any point bounded by the $95$th percentile curve for that value of $n$.  However, a new color is not assigned if a point has already been assigned a color for some lower value of $n$, to be consistent with the approximation given in \cref{sec:methods:safetymargins} (note the $\forall$ statement); such situations occur when $b_{\beta}^{*}\left(p(\cdot), n\right)$ is not a monotonically-increasing function of $n$, for some $p(\cdot)$.  We have not observed such non-monotonicity for Beamrider, but have observed it for Pong, as we describe in \cref{sec:app:results}.

A key result is that despite the noisy, weak relationship between the proxy criticality metric and true criticality, there is a much clearer relationship between the proxy metric and safety margins, as indicated by the shape of the region boundaries in the heatmaps of \cref{fig:results:safetymargins:beamrider}: these margins often decrease as proxy criticality increases.  Importantly, safety margins provide assurances even in the presence of false negative errors in the proxy criticality metric (where the value of the metric is low, but true criticality is high, e.g., as seen in \cref{fig:results:kde:a3c:pong:excerpt}): the majority of true criticality values (roughly $\beta$ of them) are still less than $b_{\beta}^{*}\left(p(\cdot), n\right)$. A drawback is that with a high false negative error rate, safety margins become overly conservative (small), making a relatively safe situation appear unsafe. Too many of these errors indicate that a more accurate proxy criticality metric should be used.

False positive errors (where proxy criticality is high, though true criticality is low) tend to have minimal impact on safety margins, as a result of the $95$th percentile and the monotonicity constraint. This can be seen, e.g., in \cref{fig:results:kde:a3c:beamrider} for $n\leq 4$, where a large number of samples demonstrate a true criticality of about $0$, even at the highest proxy criticality values. Even with all of those false positives, the safety margin in \cref{fig:results:safetymargins:beamrider} is well behaved as a result of the $95$th percentile, keeping the probability of a loss in reward greater than $\zeta$ to roughly $\beta$.

\begin{figure*}[t]
    \centering
    \includegraphics[clip, width=\linewidth]
    {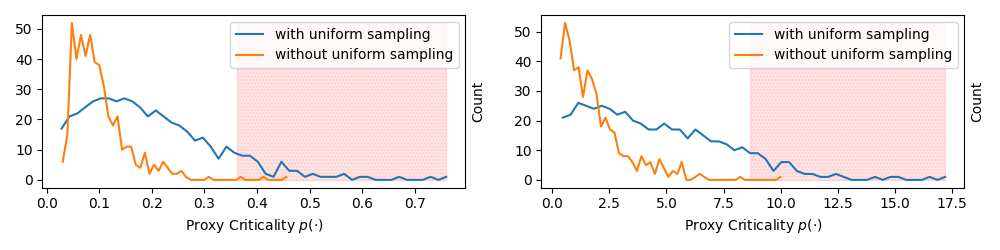}
    \caption{Histograms of metric values $p_{m}(t_m)$ for Beamrider, with vs. without uniform sampling, for APE-X (left) and A3C (right); 50 equal-sized bins were used.  The shaded region (on the right of each plot) corresponds to the top $5\%$ of $p_{m}(t_m)$ values, which were removed.}
    \label{fig:results:metrichistogram:beamrider}
\end{figure*}

The distributions of proxy criticalities from which the safety margins for Beamrider were generated are shown in \cref{fig:results:metrichistogram:beamrider}. This figure indicates the truncated $5\%$ of data tuple values via shading, and supports the necessity of uniform sampling to ensure adequate coverage of high proxy criticalities.  The distribution additionally yields important information with regards to false positives and their impact on safety margins: for a deployed agent, the natural distribution (``without uniform sampling'') is the one that is expected to be observed. In this distribution, high criticalities are rare -- indicating that any negative impact from false positives will be minor.

\subsection{Statistical Validation} \label{sec:statisticalvalidation}

The safety margins were generated using $95$th percentile curves computed via KDE plots; however, as discussed in \cref{analysis:safetymargins}, the estimated margins sampled from a limited set of observations will not hold exactly $95\%$ of the time in practice.  To determine whether the percentile error, $\epsilon_{percentile}$, is significant for a number of samples, we perform a cross-validation, where we partition the $M = 1,000$ data tuples into a ``training set'' and a ``test set''.  For the training set, we selected the first $400$ data tuples generated without uniform sampling and the first $400$ data tuples generated with uniform sampling; subsequently, as before, we removed $5\%$ of the tuples with the highest proxy criticality values, resulting in a training set size of $760$.  From this training set, we generated 2d KDE plots and percentile curves, using the approach described earlier; no attempt was made to make the curves monotonic (which is a postprocessing step that would otherwise be applied before generating safety margin heatmaps).  Then, for each of the $200$ data tuples in the ``test set'', and for each value $n$, we determine whether $c_{m}^{*}(t_{m}, n; \pi) \leq b_{\beta}^{*}\left( p(\cdot), n \right)$, where $c_{m}^{*}(t_{m}, n; \pi)$ is obtained from the test set; $b_{\beta}^{*}\left( p(\cdot), n \right)$ is computed over the training set; and $p(\cdot)$ is chosen as proxy criticality bin value (in the KDE plot) that is closest to $p_{m}(t_{m})$.  If the inequality holds, then the safety margin has succeeded for the given tuple and $n$ value.  Finally, we estimate the percentile error $\epsilon_{percentile}$ by subtracting the observed success rate from the expected $\beta = 0.95$.

In \cref{cai:tab:results:percentiles}, we provide the estimated percentile errors for different environment-algorithm combinations, at different values $n$.  In many cases, we observe a negative percentile error, which indicates a higher than expected (above $0.95$) success rate.  One possible explanation for this is that the KDE bandwidths are not sufficiently tuned, across the true and proxy criticality dimensions. In particular, we observe that the natural distribution skews heavily toward low proxy criticality values, as indicated in \cref{fig:results:metrichistogram:beamrider,fig:results:metrichistogram:pong}, with few proxy criticality values below (i.e., to the left) of the peak of the distribution, but many values above (to the right) of it.  For some given proxy criticality value/bin near the peak, higher proxy criticality values are thus oversampled, and they tend to have higher corresponding true criticality, potentially leading to overestimated percentiles; the effect becomes more pronounced with a higher proxy criticality bandwidth.  As we describe in \cref{sec:app:setup}, a single bandwidth value is used for any horizontal or vertical bin within the KDE plot, regardless of the sampling density in/near that bin.

As mentioned in \cref{analysis:safetymargins}, the computed percentile curves are also potentially subject to sampling error in the true criticality estimates, but this can be addressed by setting the true criticality (vertical axis) bandwidth to at least $\hat{\epsilon}_{sampling} / t_{\alpha}$.  In our experiments, we used $\hat{\epsilon}_{sampling} = 0.2$, computed at a confidence level $\alpha = 0.95$.  Since $t_{\alpha=0.95}$ approaches $1.96$ when the sample size $N$ is large (with higher values at lower sample sizes), $\hat{\epsilon}_{sampling} / t_{\alpha} \leq 0.102$.  Across all of the environment-algorithm-$n$ combinations, the lowest vertical bandwidth is $0.103$, for Beamrider, A3C, and $n = 1$; note that vertical bandwidth tends to increase with higher values of $n$.  Thus, the bandwidth used for KDE calculations is sufficiently high to absorb criticality errors.

\begin{table}
\centering
\caption{Statistical validation of percentiles}
\label{cai:tab:results:percentiles}
\begin{tabular}{|c|c|c|c|}
\hline
\multicolumn{1}{|l|}{environment} & \multicolumn{1}{l|}{algorithm} & n  & \begin{tabular}[c]{@{}c@{}}estimated\\ $\epsilon_{percentile}$\end{tabular} \\ \hline
\multirow{12}{*}{Pong}            & \multirow{6}{*}{APE-X}         & 1  & -0.010                                                                      \\ \cline{3-4} 
                                  &                                & 2  & -0.015                                                                      \\ \cline{3-4} 
                                  &                                & 4  & 0.000                                                                       \\ \cline{3-4} 
                                  &                                & 8  & 0.020                                                                       \\ \cline{3-4} 
                                  &                                & 16 & 0.005                                                                       \\ \cline{3-4} 
                                  &                                & 32 & -0.025                                                                      \\ \cline{2-4} 
                                  & \multirow{6}{*}{A3C}           & 1  & -0.035                                                                      \\ \cline{3-4} 
                                  &                                & 2  & -0.020                                                                      \\ \cline{3-4} 
                                  &                                & 4  & -0.040                                                                      \\ \cline{3-4} 
                                  &                                & 8  & -0.045                                                                      \\ \cline{3-4} 
                                  &                                & 16 & -0.050                                                                      \\ \cline{3-4} 
                                  &                                & 32 & -0.045                                                                      \\ \hline
\multirow{12}{*}{Beamrider}       & \multirow{6}{*}{APE-X}         & 1  & 0.005                                                                       \\ \cline{3-4} 
                                  &                                & 2  & 0.000                                                                       \\ \cline{3-4} 
                                  &                                & 4  & 0.010                                                                       \\ \cline{3-4} 
                                  &                                & 8  & -0.005                                                                      \\ \cline{3-4} 
                                  &                                & 16 & -0.005                                                                      \\ \cline{3-4} 
                                  &                                & 32 & 0.005                                                                       \\ \cline{2-4} 
                                  & \multirow{6}{*}{A3C}           & 1  & -0.010                                                                      \\ \cline{3-4} 
                                  &                                & 2  & 0.000                                                                       \\ \cline{3-4} 
                                  &                                & 4  & -0.025                                                                      \\ \cline{3-4} 
                                  &                                & 8  & 0.005                                                                       \\ \cline{3-4} 
                                  &                                & 16 & -0.005                                                                      \\ \cline{3-4} 
                                  &                                & 32 & 0.000                                                                       \\ \hline
\end{tabular}
\end{table}

\subsection{Domain-Specific Validation} \label{sec:domainspecificvalidation}

\begin{table}
    \centering
    \caption{Predictive capabilities of safety margins}
    \label{cai:tab:results:safety}
    \begin{tabular}{|c|c|c|c|}
    \hline
    $\zeta$ & algorithm & steps before death & safety margin \\\hline
    \multirow{6}{*}{0.5} & \multirow{4}{*}{APE-X} & 1 & $0.00 \pm 0.00$ \\
    & & 2 & $0.50 \pm 0.87$ \\
    & & 4 & $0.99 \pm 0.73$ \\
    & & average & $0.97 \pm 0.86$ \\\cline{2-4}
    & \multirow{4}{*}{A3C} & 1 & $0.53 \pm 0.50$ \\
    & & 2 & $1.00 \pm 0.00$ \\
    & & 4 & $0.82 \pm 0.38$ \\
    & & average & $3.09 \pm 2.68$ \\\hline
    \multirow{6}{*}{1.0} & \multirow{4}{*}{APE-X} & 1 & $2.00 \pm 0.00$ \\
    & & 2 & $2.50 \pm 0.87$ \\
    & & 4 & $4.50 \pm 2.23$ \\
    & & average & $4.72 \pm 2.50$ \\\cline{2-4}
    & \multirow{4}{*}{A3C} & 1 & $2.48 \pm 1.75$ \\
    & & 2 & $4.00 \pm 0.00$ \\
    & & 4 & $3.42 \pm 2.66$ \\
    & & average & $9.20 \pm 5.24$ \\
    \hline
    \end{tabular}
\end{table}

For safety margins to have real-world value, they must have a significant negative correlation with the benefits of deeper analysis or intervention in a given situation; i.e., a lower safety margin should indicate that human oversight is more likely to be needed, because a highly undesirable state may arise in the near future.  In the case of environments like Beamrider, a particularly undesirable state is one in which the agent is destroyed. However, the agent's own actions led to this destruction, meaning that true criticality will be low -- recall that a high criticality only results when perturbations to the agent's policy, $\pi$, result in worse performance (\cref{sec:methods:definition}).  Fortunately, for a proxy criticality metric such as the one used here \cite{lin2017}, the agent's policy is learned as a statistical expectation over observed scenarios. That is, the catastrophic action was chosen as a result of the agent perceiving it to be beneficial in similar scenarios. When those similar scenarios would have high true criticality, as the agent's policy is closer to optimal for them, we would typically expect the proxy metric to be high, resulting in low safety margins.

To confirm the above hypothesis, that safety margins will be low even in scenarios where the agent performs suboptimally, we performed an experiment where, for each of $100$ episodes of Beamrider, we computed the safety margins $s^*(p(\cdot), \zeta; \pi, S)$ for different values $\zeta$ at $1$, $2$ and $4$ time steps before the agent's death, as well as the average safety margin per time step of the episode.  (We did not perform a similar experiment for Pong, since it is rare for the models to lose points.)  In \cref{cai:tab:results:safety} (which first appeared in \cite{grushin2023}), we provide the means and standard deviations (before and after the $\pm$ sign, respectively) of these safety margin values across the $100$ episodes.  We observe that safety margins tend to decrease as the agent nears its own destruction, and are much lower than the average safety margins (over all times in all episodes). This indicates the ability of our defined safety margins to correctly identify when the agent would benefit from human oversight.  We further note that for APE-X, in $22\%$ of cases, the proxy criticality value just before the agent's death is within the top $5\%$ of proxy criticality values sampled across all times in all episodes, and for A3C, the corresponding proportion is $47\%$. This suggests that, even if the human intervenes only when proxy criticality is very high, safety can potentially be improved with a limited amount of oversight; further improvements might be yielded with more accurate proxy metrics.

\subsection{Discussion} \label{sec:discussion}

We have presented and demonstrated an approach for converting proxy criticality metrics, which can suffer from poor accuracy and are expressed in units with no clear connection to application performance, into a set of statistically accurate and intuitive safety margins.  At the core of our approach is a definition of true criticality.  While this definition assumes an accurate policy $\pi$, we demonstrated that meaningful safety margins can be generated even when the policy is incorrect (\cref{sec:domainspecificvalidation}).  
In critical situations with a bad policy, true criticality is low according to the definition, but proxy criticality is still typically high (\cref{sec:domainspecificvalidation}).  
Though this might be viewed as a false positive, it overcomes the limitation of the definition in a way that allows for safety margins to be useful for agents exhibiting suboptimal performance.

Given the definition, we developed an algorithm that tractably computes an approximation of true criticality, $c^{*}(t, n; \pi)$, by repeatedly perturbing some number of actions $n$, and measuring the mean reward reduction. 
This algorithm differs from those used in existing work on proxy criticality metric evaluation \cite{lin2017,sun2020,guo2021,kumar2021} in that the actions exist within a local time interval $(t, t+1, ..., t+n-1)$ (rather than being scattered in time), allowing for criticality to be approximated locally to that time interval. Additionally, multiple trials are performed, with different random choices made when perturbing actions (in order to control sampling error), rather than perturbing the chosen set of actions only once.

To collect the statistics required to compute safety margins, pairs of corresponding approximate true criticality and proxy criticality metric values, $c_{m}^{*}(t_{m}, n; \pi)$ and $p_{m}(t_{m}, n; \pi)$, respectively, are collected across multiple episodes $m$, from both a natural distribution and an approximate uniform distribution with respect to the proxy criticality values (\cref{sec:methods:collection}).
A key feature of this approach is that it selects time steps $t_{m}$ with a broad range of criticality values, both high and low, which captures both false negative and false positive errors in the criticality metric.  By contrast, the aforementioned existing approaches \cite{lin2017,sun2020,guo2021,kumar2021} focus only on time steps that have high criticality values, and can result in missed false negative errors.  We have demonstrated how, by representing the generated proxy and true criticality data via 2d KDE plots and computing high percentiles on true criticality (as functions of proxy criticality), we can generate approximate safety margins for reinforcement learning models.

\section{Conclusion and Future Work}

We introduced safety margins, a bound on the number of random actions tolerated by an autonomous controller before the performance impact exceeds some threshold $\zeta$. It was shown that safety margins can be computed based on: (1) a definition of true criticality as the expected drop in reward when an agent's policy is perturbed to emit uniform random actions for $n$ time steps; and (2) the availability of a proxy metric that roughly monotonically corresponds to true criticality, such as the agent's perceived difference in expected reward between the best and worst available actions. An algorithm was introduced for computing true criticality accurately but slowly; in contrast, the proxy metric can be computed almost instantaneously. Kernel density estimates were used to analyze the relationship between the true and proxy criticalities for different numbers of perturbed time steps, $n$, and multiple such kernel density estimates can be compiled into a single, intuitive heatmap for a model (e.g., \cref{fig:results:safetymargins:beamrider}). We demonstrated that, in an Atari environment where the agent can lose, safety margins decrease as the agent nears a loss, with 47\% of losses being within the 5\% smallest safety margins. This was achieved with a very simple proxy metric, with limited expressiveness \cite{lin2017}. We believe that these methods can find practical application in two different ways. First, aiding in post-hoc analysis of autonomous agents: by focusing only on decisions that were below some threshold of safety margin, the amount of analysis work can be significantly reduced in a way that does not suffer from false negatives like prior work. Second, aiding in the active management of multiple autonomous controllers: safety margins can be used to draw human attention to autonomous controllers only when it is likely to be beneficial, potentially enabling less human effort to avoid a greater number of catastrophic autonomous decisions.

\subsection{Future Work} \label{sec:future}

We expect the approach for safety margins -- having a computable definition of true criticality and statistically binding it to proxy criticality metrics -- to support our use cases of better autonomous agent debugging and active oversight of multiple autonomous agents. 
That said, aspects of the methodology in this paper could likely be improved by future work.
\vspace*{0.5em}

\noindent From a methodology perspective:
\begin{enumerate}
    \item An exploration of additional proxy metrics and their utility for different reinforcement learning algorithms. This could include the development of new proxy metrics that leverage $n$ via, e.g., world models \cite{ha2018,schrittwieser2020}, or safety margin-specific proxy metrics that are learned from a large set of true criticalities and corresponding neural network feature values.
    \item An investigation into whether an alternative definition of true criticality could avoid the confusing situation where proxy metric is high and true criticality is low, not because of a false positive, but because of a suboptimal agent policy $\pi$.
    \item A detailed study of the statistical properties resulting from safety margin confidences higher than 95\%.
\end{enumerate}

% ==============================================
% Manually balance; basically, specify `\lastcolumnfix` to happen in the first column of the last page. 
\newcommand\lastcolumnfix{\enlargethispage{-1.25cm}}
%\IEEEtriggercmd{\lastcolumnfix}
\lastcolumnfix
%\IEEEtriggeratref{13} 
% ==============================================

\noindent From a use case perspective:
\begin{enumerate}
    \item A user study evaluating the utility of safety margins for human teams operating autonomous agents.
    \item A study of the accuracy of safety margins when transferred between a simulator and the real world, and the impacts of potential methodological improvements to that transfer (which could include dynamics randomization within the simulator \cite{peng2017}).
    \item Research into the relationship between safety margins and network training stability, especially for techniques designed to support improving the learned policy based on a human analyst's annotations, via techniques such as \cite{santurkar2021editing}.
    \item A study of the relationship between safety margins and adversarial attacks against these networks; that is, do adversarial attacks decrease safety margins, or is there a way to increase the proxy criticality attached to instances of such attacks?
\end{enumerate}

Our hope is that with these research directions, automatically-generated safety margins will eventually allow for safer operations when deploying trained reinforcement learning models in real-world applications.

\section*{Acknowledgments}

This material is based upon work supported by the Air Force Research Laboratory (AFRL) under Contract No. FA8750-22-C-1002.  This paper was PA approved under case number AFRL-2023-5871.  We would like to thank Adam Karvonen and Kaitlyn Laurie for additional literature review.
\vspace{-0.15em}

\bibliographystyle{sty/ieee/IEEEtran-nomonth}
\bibliography{sty/ieee/IEEEabrv,references}

\begin{thebibliography}{10}
\def\url#1{}
\csname url@samestyle\endcsname
\providecommand{\newblock}{\relax}
\providecommand{\bibinfo}[2]{#2}
\providecommand{\BIBentrySTDinterwordspacing}{\spaceskip=0pt\relax}
\providecommand{\BIBentryALTinterwordstretchfactor}{4}
\providecommand{\BIBentryALTinterwordspacing}{\spaceskip=\fontdimen2\font plus
\BIBentryALTinterwordstretchfactor\fontdimen3\font minus \fontdimen4\font\relax}
\providecommand{\BIBforeignlanguage}[2]{{%
\expandafter\ifx\csname l@#1\endcsname\relax
\typeout{** WARNING: IEEEtran.bst: No hyphenation pattern has been}%
\typeout{** loaded for the language `#1'. Using the pattern for}%
\typeout{** the default language instead.}%
\else
\language=\csname l@#1\endcsname
\fi
#2}}
\providecommand{\BIBdecl}{\relax}
\BIBdecl

\bibitem{lin2017}
Y.-C. {Lin}, Z.-W. {Hong} \emph{et~al.}, ``{Tactics of Adversarial Attack on Deep Reinforcement Learning Agents},'' \emph{arXiv e-prints arXiv:1703.06748}, 2017.

\bibitem{ribeiro2016}
M.~{Tulio Ribeiro}, S.~{Singh} \emph{et~al.}, ``{``Why Should I Trust You?'': Explaining the Predictions of Any Classifier},'' \emph{arXiv e-prints arXiv:1602.04938}, 2016.

\bibitem{woods2019}
W.~{Woods}, J.~{Chen} \emph{et~al.}, ``{Adversarial Explanations for Understanding Image Classification Decisions and Improved Neural Network Robustness},'' \emph{arXiv e-prints arXiv:1906.02896}, 2019.

\bibitem{huber2023}
T.~{Huber}, M.~{Demmler} \emph{et~al.}, ``{GANterfactual-RL: Understanding Reinforcement Learning Agents' Strategies through Visual Counterfactual Explanations},'' \emph{arXiv e-prints arXiv:2302.12689}, 2023.

\bibitem{quan2019}
A.~Quan, C.~Herrmann \emph{et~al.}, ``Project vulture: A prototype for using drones in search and rescue operations,'' in \emph{2019 15th International Conference on Distributed Computing in Sensor Systems (DCOSS)}, pp. 619--624, 2019.

\bibitem{huang2018}
S.~{Huang}, K.~{Bhatia} \emph{et~al.}, ``{Establishing Appropriate Trust via Critical States},'' \emph{arXiv e-prints arXiv:1810.08174}, 2018.

\bibitem{spielberg2018}
Y.~{Spielberg} and A.~{Azaria}, ``{The Concept of Criticality in Reinforcement Learning},'' \emph{arXiv e-prints arXiv:1810.07254}, 2018.

\bibitem{sun2020}
J.~{Sun}, T.~{Zhang} \emph{et~al.}, ``{Stealthy and Efficient Adversarial Attacks against Deep Reinforcement Learning},'' \emph{arXiv e-prints arXiv:2005.07099}, 2020.

\bibitem{guo2021}
\BIBentryALTinterwordspacing
W.~Guo, X.~Wu \emph{et~al.}, ``Edge: Explaining deep reinforcement learning policies,'' in \emph{Advances in Neural Information Processing Systems}, M.~Ranzato, A.~Beygelzimer \emph{et~al.}, Eds., vol.~34, pp. 12\,222--12\,236.\hskip 1em plus 0.5em minus 0.4em\relax Curran Associates, Inc., 2021.  \url{https://proceedings.neurips.cc/paper/2021/file/65c89f5a9501a04c073b354f03791b1f-Paper.pdf}
\BIBentrySTDinterwordspacing

\bibitem{kumar2021}
R.~Praveen~Kumar, I.~Niranjan~Kumar \emph{et~al.}, ``Critical state detection for adversarial attacks in deep reinforcement learning,'' in \emph{2021 20th IEEE International Conference on Machine Learning and Applications (ICMLA)}, pp. 1761--1766, 2021.

\bibitem{xu2021}
L.~Xu, F.~Wu \emph{et~al.}, ``Criticality-guided deep reinforcement learning for motion planning,'' in \emph{2021 China Automation Congress (CAC)}, pp. 3378--3383, 2021.

\bibitem{spielberg2022}
Y.~{Spielberg} and A.~{Azaria}, ``Proceedings of the 44th annual conference of the cognitive science society,'' in \emph{{Criticality-Based Advice in Reinforcement Learning}}, vol.~44, 2022.

\bibitem{horgan2018}
D.~{Horgan}, J.~{Quan} \emph{et~al.}, ``{Distributed Prioritized Experience Replay},'' \emph{arXiv e-prints arXiv:1803.00933}, 2018.

\bibitem{mnih2016}
V.~{Mnih}, A.~{Puigdom{\`e}nech Badia} \emph{et~al.}, ``{Asynchronous Methods for Deep Reinforcement Learning},'' \emph{arXiv e-prints arXiv:1602.01783}, 2016.

\bibitem{grushin2023}
A.~{Grushin}, W.~{Woods} \emph{et~al.}, ``Safety margins in reinforcement learning,'' in \emph{2023 IEEE Conference on Artificial Intelligence (2023 IEEE CAI)}, pp. 42--43, 2023.

\bibitem{driels2004}
M.~Driels and Y.~Shin, ``Determining the number of iterations for monte carlo simulations of weapon effectiveness,'' Department of Mechanical \& Astronautical Engineering, Naval Postgraduate School, 700 Dryer Rd, Monterey, CA 93943-5000, Tech. Rep. NPS-MAE-04-005, 2004.

\bibitem{ha2018}
D.~{Ha} and J.~{Schmidhuber}, ``{World Models},'' \emph{arXiv e-prints arXiv:1803.10122}, 2018.

\bibitem{schrittwieser2020}
J.~{Schrittwieser}, I.~{Antonoglou} \emph{et~al.}, ``{Mastering Atari, Go, chess and shogi by planning with a learned model},'' \emph{arXiv e-prints arXiv:1911.08265}, 2020.

\bibitem{peng2017}
X.~B. {Peng}, M.~{Andrychowicz} \emph{et~al.}, ``{Sim-to-Real Transfer of Robotic Control with Dynamics Randomization},'' \emph{arXiv e-prints arXiv:1710.06537}, 2017.

\bibitem{santurkar2021editing}
S.~Santurkar, D.~Tsipras \emph{et~al.}, ``Editing a classifier by rewriting its prediction rules,'' \emph{Advances in Neural Information Processing Systems}, vol.~34, pp. 23\,359--23\,373, 2021.

\bibitem{scott1992}
\BIBentryALTinterwordspacing
D.~Scott, \emph{Multivariate Density Estimation: Theory, Practice, and Visualization}, ser. A Wiley-interscience publication.\hskip 1em plus 0.5em minus 0.4em\relax Wiley, 1992.  \url{https://books.google.com/books?id=7crCUS_F2ocC}
\BIBentrySTDinterwordspacing

\bibitem{ialongo2018}
C.~Ialongo, ``\BIBforeignlanguage{en}{Confidence interval for quantiles and percentiles},'' \emph{\BIBforeignlanguage{en}{Biochem Med (Zagreb)}}, vol.~29, no.~1, p. 010101, 2018.

\end{thebibliography}

\clearpage
\appendix

In the following, we begin by presenting several alternative definitions of criticality, and explain why they were not favored over the definition given in \cref{sec:methods:definition}.  We then provide the details of our experimental procedures, and present some results that were not included in \cref{sec:results}.  Finally, we provide a theoretical analysis of the relationship between percentile error $\epsilon_{percentile}$ (\cref{analysis:safetymargins}) and the number of tuples $M$ used.

\subsection{Alternative Definitions of True Criticality} \label{sec:app:alternativedefinitions}

A formal definition of true criticality serves as the foundation of the approach that we have presented in this paper.  Of course, various other definitions could be proposed; however, we found that the alternatives that we have considered are less suitable for our purposes.  Example alternative definitions include the following:

\begin{itemize}
    \item Rather than defining $\mathbb{E}_{a \sim \pi}\left[R_{\gamma}\right]$ in terms of the agent's policy $\pi$, we could define it in terms of an {\it optimal} policy $\pi^{*}$, which selects the best action possible at every time step.  This definition has the advantage of being independent of the agent's policy $\pi$ (and the actions that it prescribes), with criticality depending only on the state of the environment.  However, the definition is impractical, since approximating criticality would then require an optimal or near-optimal policy to be found, which itself can be an intractable problem.  While our existing definition can, in theory, be problematic in situations where the agent's policy is incorrect, we may still be able to effectively approximate safety margins in such situations (as we demonstrated in \cref{sec:domainspecificvalidation}), or adjust the definition in a more practical way (as we proposed in \cref{sec:discussion}).
    \item Another approach is to define criticality as the expected variance (or standard deviation) in the total discounted reward if $\pi'(t, n)$ is followed throughout the episode, i.e., with random action choices being made at time steps $t, t+1, ..., t+n-1$ and the normal policy $\pi$ followed at other time steps; in other words, $c(t, n; \pi) = \mathrm{Var}_{a \sim \pi'(t, n)}\left[R_{\gamma}\right]$; note that the unperturbed reward $\mathbb{E}_{a \sim \pi}\left[R_{\gamma}\right]$ is not part of the definition.  We note that a related definition was given in \cite{spielberg2018}, with criticality being proportional to the variance in the optimal Q function; a variance-based definition was also suggested in \cite{spielberg2022}.  A drawback of variance-based definitions is that variance $\mathrm{Var}_{a \sim \pi'(t, n)}\left[R_{\gamma}\right]$ does not fully capture the typical consequences of disregarding the policy.  For example, suppose that there are $21$ action choices, and that the choice suggested by the policy will lead to a reward of $100$, while the remaining $20$ choices will lead to a reward of $-100$; not following the policy will result in a reward reduction of $200$, while the variance $\mathrm{Var}_{a \sim \pi'(t, n)}\left[R_{\gamma}\right]$ across the $21$ choices is approximately $1814$.  Now, suppose that the choices have associated rewards of $-100, -90, ..., -10, 0, 10, ..., 90, 100$; in this case, replacing the policy-prescribed action (with a reward of $100$) by a random action (with a mean reward of $0$) will lead to a lower mean reward reduction of $100$, despite the higher variance of approximately $3667$.
    \item Yet another approach is to perturb an action by replacing the best action (according to the policy $\pi$) with the worst action (according to the same policy), rather than with a random action.  While this approach has been taken in adversarial settings (e.g., in \cite{lin2017}), we found that it can result in missed false negative errors: the worst action according to $\pi$ might not actually be a bad action, resulting in an underestimate for the true criticality value.
    \item We could use the definition given in \cref{sec:methods:definition}, but compute the total reward without discounting.  While in some applications, it may be more meaningful to express criticality in terms of reductions in total reward (e.g., in terms of the total number of points lost in a game), it is not a fair definition from a bounded rationality perspective, given that the agent learns to maximize total discounted reward, rather than the total raw reward.
\end{itemize}

\subsection{Experimental Setup} \label{sec:app:setup}

In the following, we describe the specifics of the safety margin generation pipeline.  The Pong and Beamrider games were implemented via the Arcade Learning Environment (ALE)\footnote{https://github.com/mgbellemare/Arcade-Learning-Environment}.  For the APE-X and A3C training algorithms, we used the implementations provided by the RLLib library for reinforcement learning\footnote{https://docs.ray.io/en/latest/rllib/index.html}, running on the Ray\footnote{https://docs.ray.io/en/latest/} framework for distributed machine learning.

For Pong, models were trained until the mean total reward (without discounting) was between $20$ and $21$ points; we note that $21$ is best mean total reward that can be obtained in Pong (the episode/game ends when one of the players has scored $21$ points, and the total reward is the difference between the number of points scored by the player, and the number of points scored by the opponent).  For Beamrider, which is a much more challenging environment with a more complex scoring system, models were trained until each one achieved a mean reward of approximately $7000$ per episode, which corresponds to an an intermediate skill level of gameplay.

We also note that in the RLLib implementation, noise is used by APE-X not only during training, but also, during deployment: when presented with the same input observation multiple times, the Q values may be slightly different each time.  However, because both the Pong and the Beamrider environments are deterministic, we estimate the total unperturbed reward over only a single trial, as prescribed in \cref{sec:methods:approximation:unperturbedreward} of \cref{alg:approximation} (i.e., the first assumption in \cref{sec:methods:approximation} is satisfied).  Furthermore, because ALE allows for the simulation state to be saved and loaded, perturbed reward is measured by running the episode from state/observation $o_{t}$ in each trial, as prescribed in \cref{sec:methods:approximation:meanperturbedreward} (the second assumption is also satisfied).

\begin{figure}[t]
    \centering
    \includegraphics[clip, trim={0.39in 1.20in 3.41in 3.82in}, width=\linewidth]
    {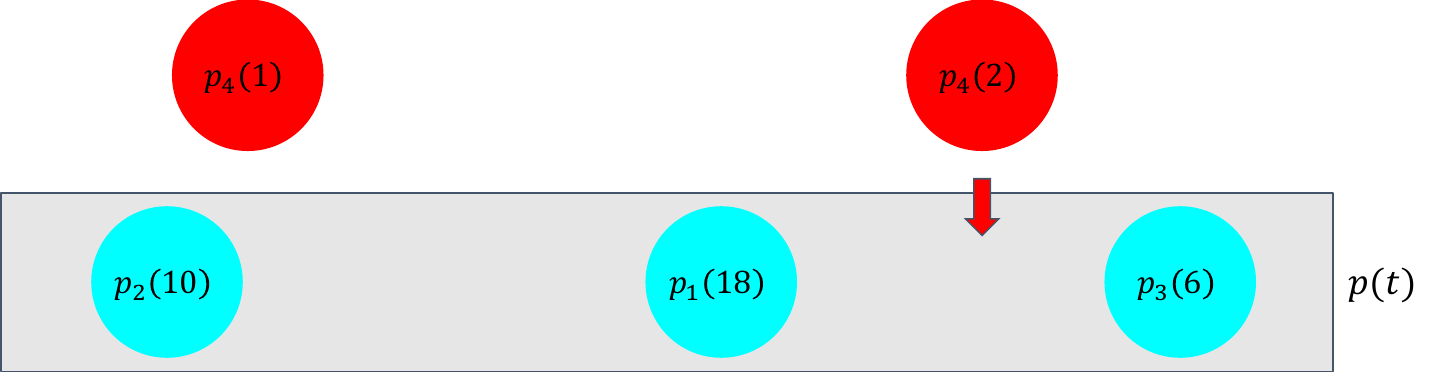}
    \caption{An illustration of the time step selection approach.  Here, three episodes have been processed so far; in these episodes, time steps $18$, $10$ and $6$ were selected, with criticality values $p_1(18)$, $p_2(10)$ and $p_3(6)$, respectively (represented via blue circles, with the horizontal position of a circle corresponding to its criticality value, where higher values appear farther to the right).  For episode four, suppose that there are only two time steps $t_{1}$ and $t_{2}$ (this is done for ease of illustration; note that for the environments that we consider, an episode will often have thousands of time steps); these have criticality values $p_4(1)$ and $p_4(2)$, and are represented via red circles.  Because $p_4(1)$ is closer than $p_4(2)$ to a criticality value for a previously-selected time step (specifically, $p_4(1)$ is only slightly higher than $p_2(10)$), time step $2$ is selected, and the criticality value $p_4(2)$ is added to the set of tracked criticality values (represented by the grey box).  Note that it is not necessary to store the previously-selected time steps, only the criticality values at these time steps.}
    \label{fig:methods:evaluation:sampling}
\end{figure}

In our experiments, we used $\hat{\epsilon}_{horizon} = 0.01$; since we used a discount factor $\gamma = 0.99$, this resulted in a horizon of $459$ time steps, when applying \cref{eqn:horizon} in \cref{sec:methods:analysis}.  For each environment, we applied the data collection procedure in \cref{alg:evaluation} for $M = 500$ episodes with the uniform sampling approach illustrated in \cref{fig:methods:evaluation:sampling}, and another $M = 500$ episodes without uniform sampling, i.e., with $t$ selected randomly and uniformly; we note that with either sampling strategy, in a given episode, the last $32$ time steps were not considered for selection.  For each episode $m$, we perturbed actions at $n \in S = \{1, 2, 4, 8, 16, 32\}$ time steps, and for each value $n$, we ran a minimum of $N_{min} = 10$ trials, with $\hat{\epsilon}_{sampling} = 0.2$ and a confidence level $\alpha = 0.95$.  With a parallelized implementation of \cref{alg:evaluation}, where multiple episodes are processed in parallel, data collection for the two environments (Pong, Beamrider) $\times$ two algorithms (APE-X, A3C) requires in the order of a week of computation time on a machine with an 8 core Intel(R) Core(TM) i7-9700K CPU @ 3.60GHz and a single ZOTAC GeForce RTX 2080 TI GPU with 4352 CUDA Cores and 11GB of GDDR6 VRAM.  However, once data is collected, safety margin generation can be performed in the order of seconds.

To generate the safety margins, we first produce 2d KDE plots, over a grid of $200 \times 200$ bins, with a Gaussian kernel, where the bandwidth (standard deviation) $H_{p(\cdot)}$ along the horizontal (proxy criticality) axis is computed as follows:
\begin{align} \label{eqn:bandwidth}
H_{p(\cdot)} &= \mathrm{stdev}^*\left[p_{m}(t_{m})\right] M^{-\frac{1}{6}}.
\end{align}

Here, $\mathrm{stdev}^*\left[p_{m}(t_{m})\right]$ is the sample standard deviation (with Bessel's correction) of the proxy criticality values across the $M$ data tuples; the exponent $-\frac{1}{6}$ is determined by Scott's Rule, which dictates that the exponent's denominator should be set to $4$ plus the number of plot dimensions ($2$ in our case) \cite{scott1992}.  For the plots, we used data from the $M = 950$ out of the $1,000$ episodes mentioned earlier, having removed the $5\%$ of data tuples with the highest proxy criticality values, since these tend to be sparse and thus very noisy.  In a given plot, the horizontal axis bounds are the minimum and maximum proxy criticality values $p_{min}(\cdot)$ and $p_{min}(\cdot)$ (respectively) found in the collected tuples.  For the vertical axis, the bounds are similarly determined by the minimum and maximum true criticality values, but with a padding of size $3 H_{c^*(\cdot)}$ added to the upper bound and subtracted from the lower bound, to ensure that almost all of the density is captured along the vertical (true criticality) axis.  Here, $H_{c^*(\cdot)}$ is the bandwidth along the vertical axis, and is computed as in \cref{eqn:bandwidth}, but with the sample standard deviation calculated over the estimated true criticality values $c_{m}^{*}(t_{m}, n; \pi)$, rather than proxy values $p_{m}(t_{m})$. The padding ensures that almost all of the density is captured along the vertical axis.  Subsequently, in any given plot, and for every proxy criticality bin, the true criticality density values are normalized, to convert them into an approximate probability distribution, and the mean, median and $95$th percentile curves are computed; contours are also added.  When generating the safety margin heatmaps, the $95$th percentile curves are first made monotonic.  Specifically, for some proxy value $p_j(\cdot)$ (at bin $j$ on the horizontal axis of the KDE plot), where the percentile $b_{\beta}^{*}\left(p_j(\cdot), n\right)$ is lower than it is for the previous proxy value $p_{j-1}(\cdot)$, we set the percentile to $b_{\beta}^{*}\left(p_{j-1}(\cdot), n\right)$.

\subsection{Additional Results}\label{sec:app:results}

\begin{figure*}[t]
    \centering
    \includegraphics[clip, width=\linewidth]
    {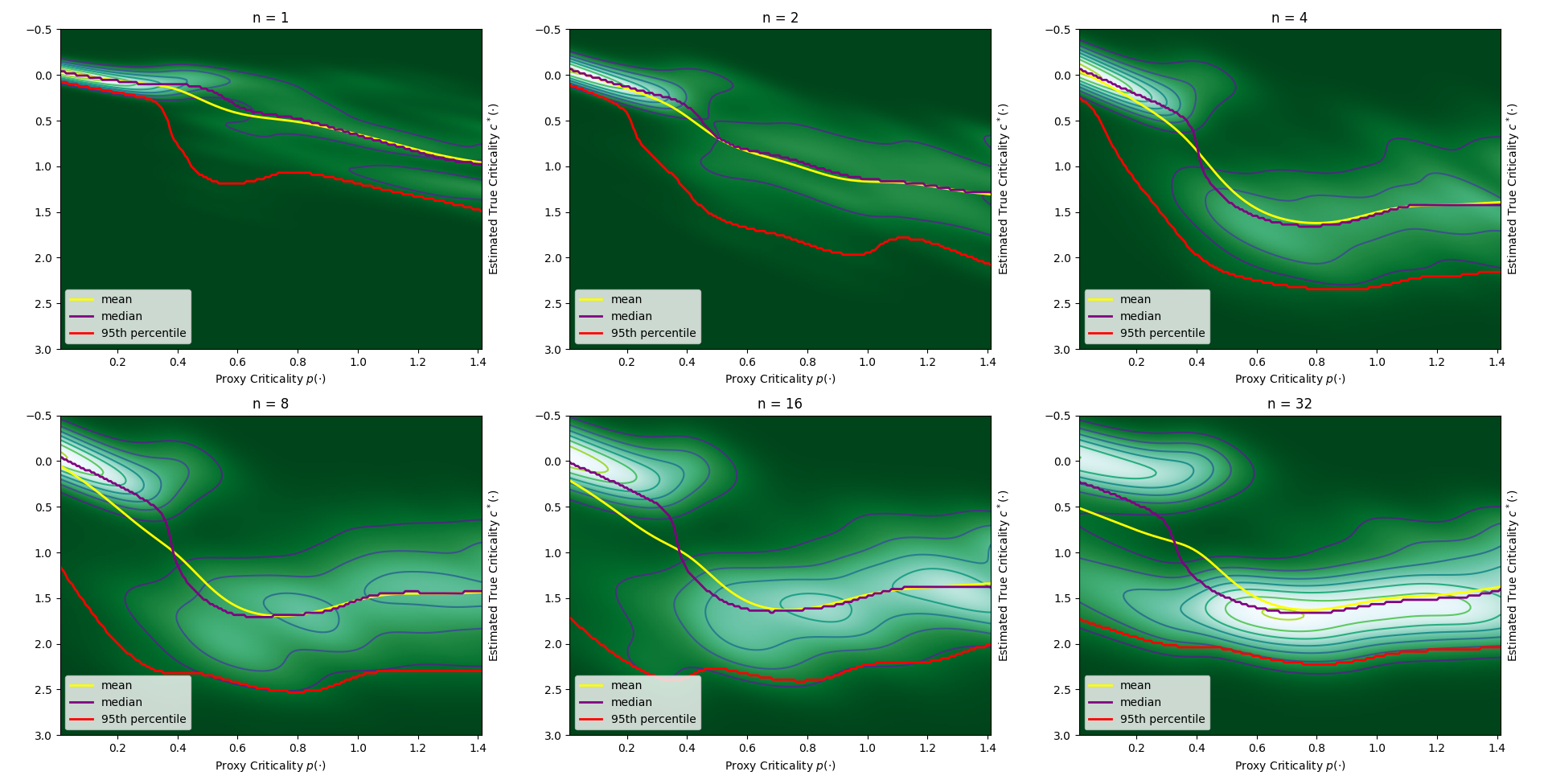}
    \caption{Normalized KDE plots capturing the relationship between proxy criticality (on the horizontal axis) and estimated true criticality (on the vertical axis), with contours and trendlines, for an APE-X model trained on Pong.}
    \label{fig:results:kde:dqn:pong}
\end{figure*}

\begin{figure*}[t]
    \centering
    \includegraphics[clip, width=\linewidth]
    {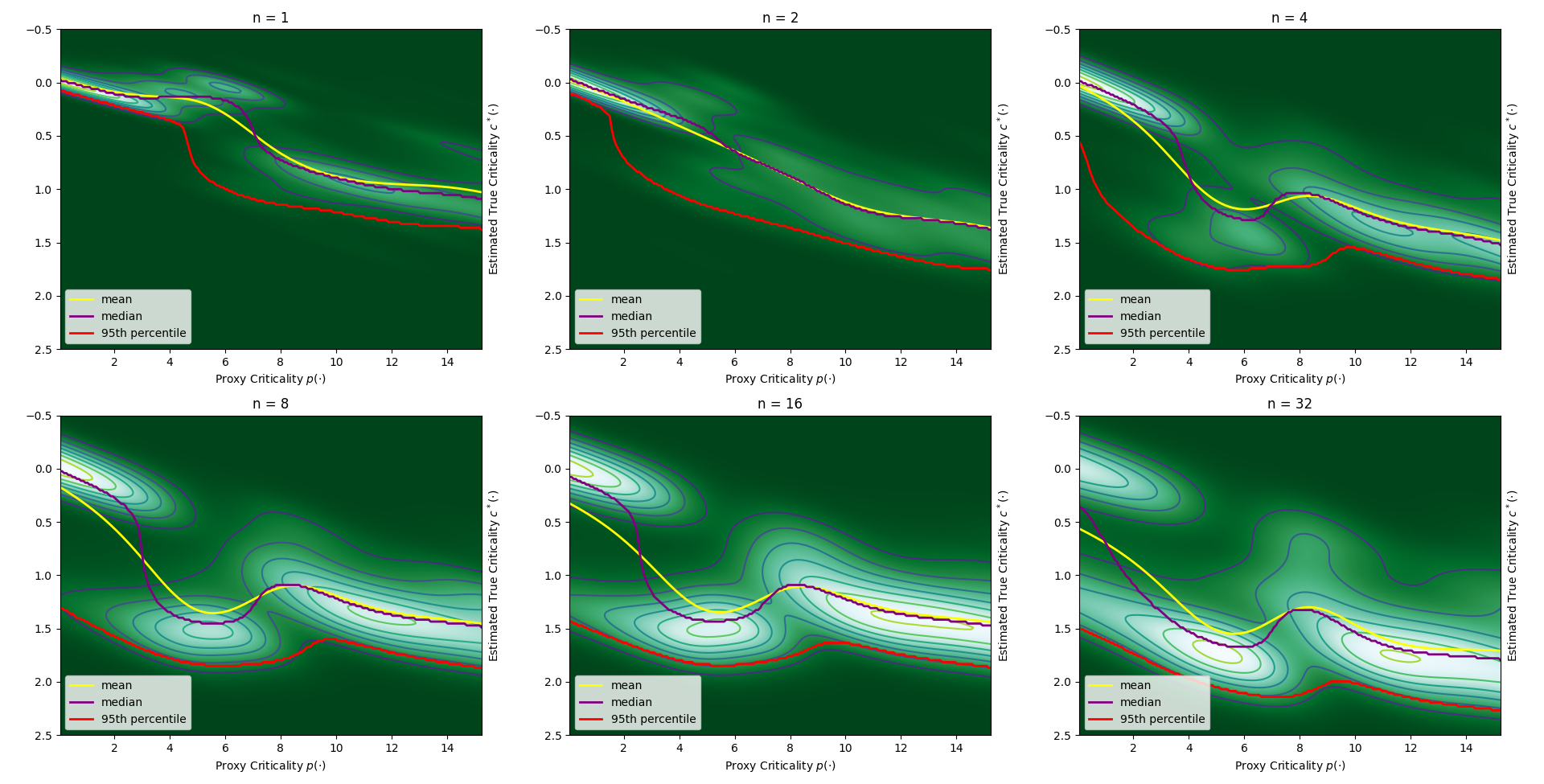}
    \caption{Normalized KDE plots capturing the relationship between proxy criticality (on the horizontal axis) and estimated true criticality (on the vertical axis), with contours and trendlines, for an A3C model trained on Pong.}
    \label{fig:results:kde:a3c:pong}
\end{figure*}

\begin{figure*}[t]
    \centering
    \includegraphics[clip, width=\linewidth]
    {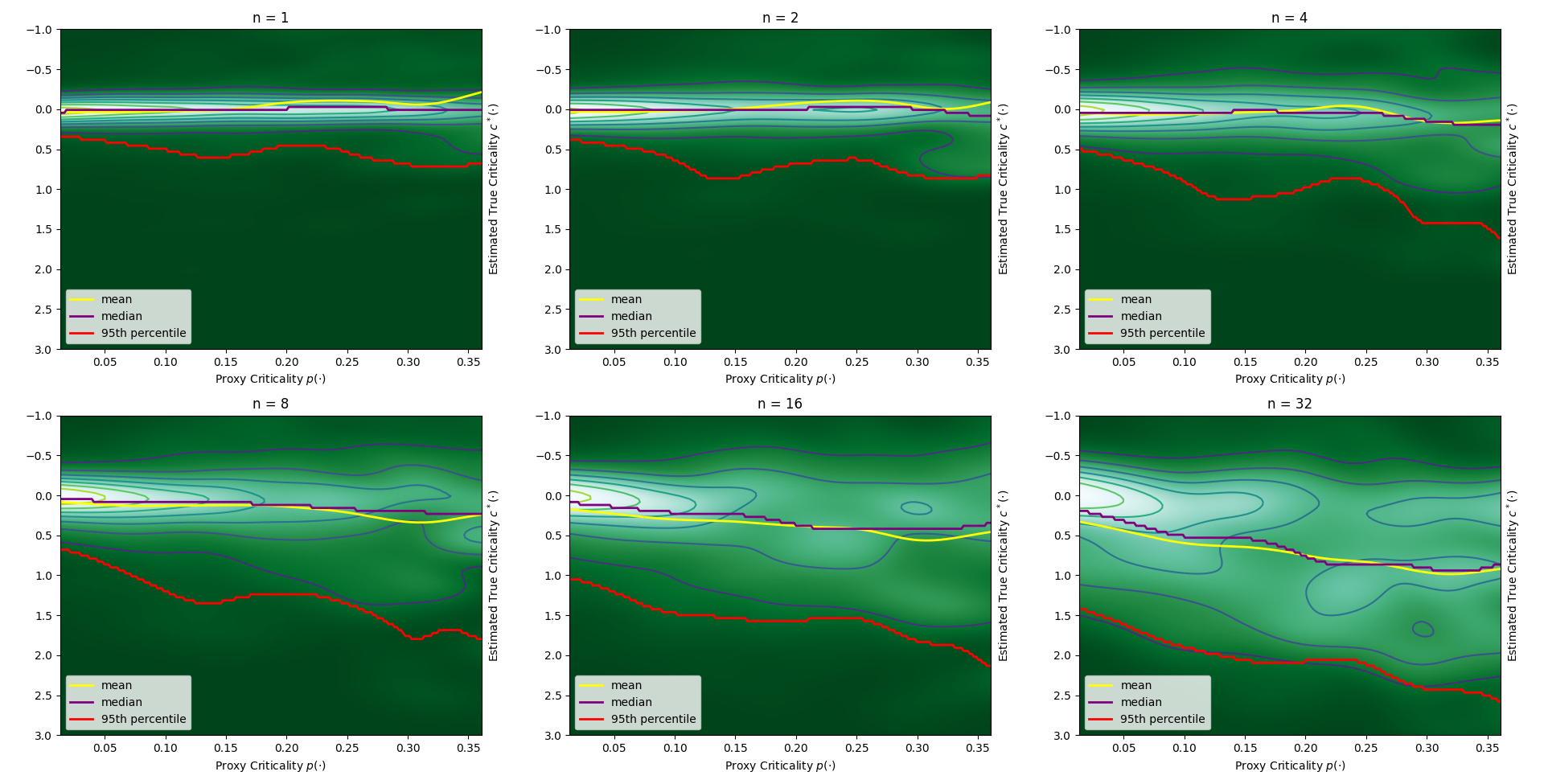}
    \caption{Normalized KDE plots capturing the relationship between proxy criticality (on the horizontal axis) and estimated true criticality (on the vertical axis), with contours and trendlines, for an APE-X model trained on Beamrider.}
    \label{fig:results:kde:dqn:beamrider}
\end{figure*}

\begin{figure*}[t]
    \centering
    \includegraphics[clip, width=\linewidth]
    {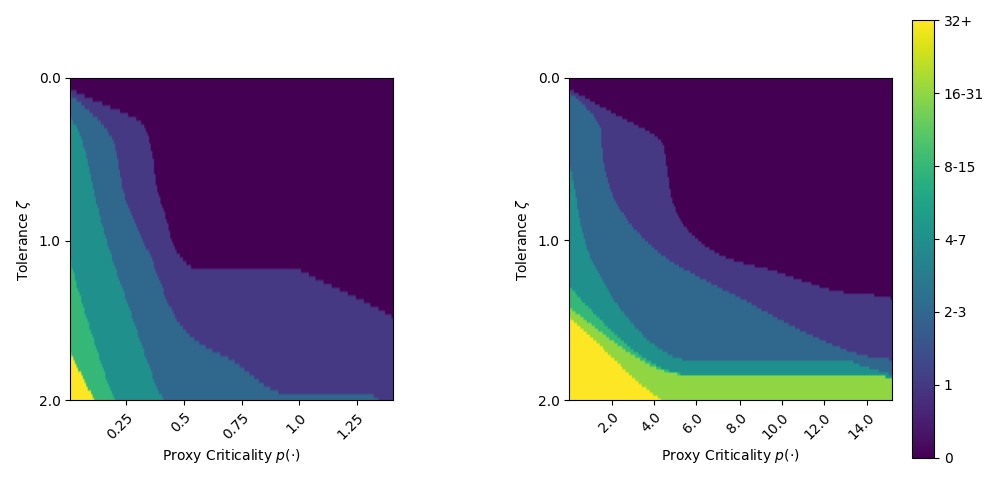}
    \caption{Heatmaps displaying safety margins for Pong in the proxy criticality metric vs. tolerance space, for APE-X (left) and A3C (right).}
\label{fig:results:safetymargins:pong}
\end{figure*}

\begin{figure*}[t]
    \centering
    \includegraphics[clip, width=\linewidth]
    {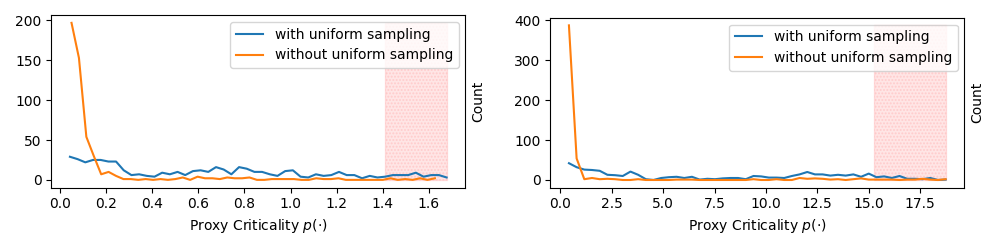}
    \caption{Histograms of metric values $p_{m}(t_m)$ for Pong, with vs. without uniform sampling, for APE-X (left) and A3C (right); 50 equal-sized bins were used.  The shaded region (on the right of each plot) corresponds to the top $5\%$ of $p_{m}(t_m)$ values, which were removed.}
    \label{fig:results:metrichistogram:pong}
\end{figure*}

In \cref{fig:results:kde:dqn:pong} and \cref{fig:results:kde:a3c:pong}, we present the KDE plots for Pong (with the APE-X and A3C algorithms, respectively), while in \cref{fig:results:kde:dqn:beamrider}, we give this plot for Beamrider with A3C.  Additionally, for Pong, \cref{fig:results:safetymargins:pong} shows safety margins, while \cref{fig:results:metrichistogram:pong} provides histograms of proxy metric values.  We note that for the Pong and APE-X combination, the mean, median and $95$th percentile curves tend to be steepest for the middle range of $n$, whereas for lower values of $n$, the impact of perturbations (i.e., true criticality) is less significant even when proxy criticality is high, and for high values of $n$, this impact becomes more significant even for lower proxy criticality.  Here, we also observe that the percentile curve for $n = 32$ is almost always less than the percentile curve for $n = 16$ (and for some higher values of $p(\cdot)$, the curve for $n = 8$); for this reason, in \cref{fig:results:safetymargins:pong} (left), there is virtually no region corresponding to a safety margin of $16-31$.  To investigate this phenomenon, we further examined the effects of action perturbations upon the APE-X model, and found that these effects can persist for a significant amount of time after the last perturbation has been made, with the model becoming temporarily ``confused''.  In particular, we observed situations where a sequence of perturbations results in the loss of more than one point, with the second point being lost after perturbations are no longer performed; interestingly, we found that in some cases, the loss of the second point is more likely when $16$, rather than $32$ perturbations are performed.  This peculiarity of the model results in a somewhat non-monotonic relationship between $n$ and $b_{\beta}^{*}\left(p(\cdot), n\right)$.  As suggested in \cref{analysis:safetymargins}, in such situations, it becomes more important to use a more dense set $S$ (e.g., all consecutive integers from $1$ through $32$, rather than just powers of $2$).  On the other hand, when A3C is used for Pong, only minor overlap exists between the percentile curves for $n = 8$ and $n = 16$, and for Beamrider (both with APE-X and with A3C), $b_{\beta}^{*}\left(p(\cdot), n\right)$ always increases monotonically with $n$.

\subsection{Bounding the Percentile Error} \label{sec:app:percentileerror}

In the following, we describe the mathematical relationship that exists between percentile error $\epsilon_{percentile}$ (defined and discussed in \cref{analysis:safetymargins}) and the sample size.  We initially make a simplifying assumption that for a specific given value $p(\cdot)$, there is some sufficiently large number $D \leq M$ of data tuples where proxy criticality is equal to $p(\cdot)$, with each tuple $m$ having some estimated true criticality value $c_{m}^{*}(t_{m}, n; \pi)$.  We then provide an approach for estimating $D$ within a KDE plot.  Finally, we apply the approach to our experimental data, present the estimated percentile error bounds, and discuss the limitations of the estimation approach.

\subsubsection{Percentile Error as a Function of Sample Size}
To quantify the percentile error $\epsilon_{percentile}$, 

We can place approximate bounds on $\epsilon_{percentile}$ by using methods described from \cite{ialongo2018}.  Consider some arbitrary $\beta^*$, as defined in \cref{analysis:safetymargins}.  Note that by the definition of percentile, for any given tuple in the sample of size $D$, the true criticality will be less than or equal to the population percentile $b_{\beta^{*}}\left(p(\cdot), n\right)$ with probability $\beta^*$.  Now, consider the total number of tuples in the sample (of size $D$) for which $c_{m}^{*}(t_{m}, n; \pi) \leq b_{\beta^{*}}\left(p(\cdot), n\right)$ holds; this number follows a binomial distribution with mean $D \beta^{*}$ and standard deviation $\sqrt{D \beta^{*}(1-\beta^{*})}$.  For a sufficiently large sample size $D$, this can be approximated by a normal distribution with the same mean and standard deviation; note that a larger sample size must be used if $\beta^{*}$ is large. Finally, since $\beta^{*}$ is a proportion, rather than a total number, we can approximate it via a normal distribution with mean $\beta^{*}$ and a standard deviation of $\frac{\sqrt{D \beta^{*}(1-\beta^{*})}}{D}$ (i.e., we divide the mean and standard deviation by the sample size $D$).  Then, with a confidence of approximately $\alpha$, the following will hold:
\begin{align} \label{eqn:percentilebound}
\epsilon_{percentile} &\leq \frac{z_{\alpha} \sqrt{D \beta^{*}(1-\beta^{*})}}{D} \leq \frac{0.5 z_{\alpha}}{\sqrt{D}}.
\end{align}

Here, $z_{\alpha}$ is the z-score, chosen such that for the standard normal distribution, $1 - \alpha$ of the values will fall above $z_{\alpha}$.  Unlike for the computation of $\epsilon_{sampling}$ in \cref{sec:methods:analysis}, we only consider one tail of the distribution when computing $\epsilon_{percentile}$, because we are primarily concerned about the case when $\beta^{*} < \beta$.  In such a case, $b_{\beta}^{*}\left(p(\cdot), n\right) < b_{\beta}\left(p(\cdot), n\right)$, i.e., the percentile is underestimated; as a result, some safety margins may be exceeded with probability greater than $1 - \beta$.

To compute the tighter bound (given in the middle of \cref{eqn:percentilebound}), we can replace $\epsilon_{percentile}$ with $\beta - \beta^{*}$, and solve for $\beta^{*}$, e.g., by applying the quadratic formula.  The looser bound, on the right side, takes advantage of the fact that since $0 \leq \beta^{*} \leq 1$, $\sqrt{\beta^{*}(1-\beta^{*})} \leq 0.5$.  A key observation here is that the bound on $\epsilon_{percentile}$ error decreases with the square root of $D$.

\subsubsection{Estimating Sample Size in a Kernel Density Plot}

In deriving the bound in \cref{eqn:percentilebound}, we made the assumption that for a given $p(\cdot)$ value, there is a sufficient number $D$ of tuples having that value.  For a real-valued proxy criticality metric, this assumption will generally not hold: given a finite number of tuples $M$, for most $p(\cdot)$ values, there will be no tuples at all, and for many of the rest, there might only be a single tuple.  To mitigate this, as mentioned earlier, we use a 2d KDE plot to smooth the data tuples.  In the following, we approximate the sample size $D$ for each each proxy criticality value $p(\cdot)$ in the KDE plot, under the assumption that the distribution of $p_{m}(t_{m})$ values among the data tuples is approximately uniform.  

\paragraph{Box kernel} For a box (uniform) kernel, the approximation is as follows:
\begin{align} \label{eqn:samplesizeapproximation:box}
D &\approx \frac{M H_{p(\cdot)}}{p_{max}(\cdot) - p_{min}(\cdot)}.
\end{align}

Here, $H_{p(\cdot)}$ is the bandwidth of the horizontal (proxy criticality) axis of the KDE plot, while $p_{min}(\cdot)$ and $p_{max}(\cdot)$ are the minimum and maximum values of $p_{m}(t_{m})$, respectively, found among the $M$ data tuples.  $D$ can be viewed as the typical number of overlapping box kernels at any given $p(\cdot)$ value, if we were to project the 2d kernels onto the horizontal axis.

\paragraph{Gaussian kernel} For a Gaussian kernel, which is used in this paper, the approximation is less trivial, because each kernel is defined over an infinite domain: at any given value $p(\cdot)$, {\it all} kernels overlap, but the kernels have uneven values (contributions).  Consider an idealized scenario, where there is an infinite number of Gaussian kernels, each with standard deviation $\sigma = H_{p(\cdot)}$, placed at proxy criticality values $p(\cdot) \in \{-\infty, \ldots, -2\delta, -\delta, 0, \delta, 2\delta, \ldots, \infty \}$, where $\delta$ is the distance between any two consecutive kernels.  The sum of the kernels at $p(\cdot) = 0$ is then as follows:
\[
    \sum_{j=-\infty}^\infty \frac{1}{\sqrt{2 \pi} \sigma} e^{-\frac{(j \delta)^{2}}{2 \sigma^2}}.
\]

Assuming that $\delta$ is sufficiently small (relative to $\sigma$), this sum can be approximated via an integral, whose value is exactly the area under a Gaussian kernel/curve (which is equal to $1$), divided by $\delta$:
\[
    \frac{1}{\delta} \sum_{j=-\infty}^\infty \frac{1}{\sqrt{2 \pi} \sigma} e^{-\frac{(j \delta)^{2}}{2 \sigma^2}} \delta \approx \frac{1}{\delta} \int_{-\infty}^\infty \frac{1}{\sqrt{2 \pi} \sigma} e^{-\frac{x^{2}}{2 \sigma^2}} dx = \frac{1}{\delta}.
\]

For $p(\cdot) = 0$, the Gaussian kernel centered on $0$ provides the largest contribution to the sum, specifically, $\frac{1}{\sqrt{2 \pi} \sigma}$.  We consider this contribution to be equivalent to $1$ unit of sample size.  (In the general case, for any $p(\cdot)$, we consider the closest kernel (with the largest value) to always contribute $1$ unit of sample size $D$.  Even if this kernel is far away from $p(\cdot)$, with its value thus being very low at $p(\cdot)$, the contribution will still become significant when the KDE plot is normalized over the true criticality (vertical) axis, as we described in \cref{sec:methods:safetymargins}.)  Any other kernel's contribution is scaled by the contribution of the closest kernel, i.e., by the largest kernel value $\frac{1}{\sqrt{2 \pi} \sigma}$.  In the above idealized scenario, the sample size at $p(\cdot) = 0$ (and at any other proxy value $p(\cdot) \in \{-\infty, \ldots, -2\delta, -\delta, 0, \delta, 2\delta, \ldots, \infty \}$) is estimated as:

\[
    D \approx \frac{\sqrt{2 \pi} \sigma}{\delta}.
\]

For other proxy values $p(\cdot) \notin \{-\infty, \ldots, -2\delta, -\delta, 0, \delta, 2\delta, \ldots, \infty \}$, the approximation error may be higher, but should still be low, given a sufficiently small $\delta$.

In the more practical case, where there is a finite number of Gaussian kernels defined over a finite interval $\left[p_{min}(\cdot), p_{max}(\cdot) \right]$, we can substitute $\frac{p_{max}(\cdot) - p_{min}(\cdot)}{M}$ for $\delta$, and $H_{p(\cdot)}$ for $\sigma$, and obtain the following:
\begin{align} \label{eqn:samplesizeapproximation:gaussian}
D &\approx \sqrt{2 \pi} \frac{M H_{p(\cdot)}}{p_{max}(\cdot) - p_{min}(\cdot)}.
\end{align}

In other words, the sample size for the Gaussian kernel is approximately equal to the sample size for the box kernel, multiplied by a constant factor of $\sqrt{2 \pi}$.  However, near the boundaries $p_{min}(\cdot)$ and $p_{max}(\cdot)$ of the interval, it may be useful to divide the approximation $D$ (as computed by \cref{eqn:samplesizeapproximation:box} or \cref{eqn:samplesizeapproximation:gaussian}) by a factor of $2$, since on one side of the boundary, there are no tuples, and thus, no kernels.

\subsubsection{Approximate Percentile Errors}

\begin{table*}
\centering
\caption{Horizontal bandwidths, estimated samples sizes, and estimated percentile error bounds}
\label{cai:tab:results:errors}
\begin{tabular}{|c|c|c|c|c|c|}
\hline
environment                & algorithm & $H_{p(\cdot)}$    & $M$ (uniform only) & $D$     & estimated $\epsilon_{percentile}$ bound \\ \hline
\multirow{2}{*}{pong}      & APE-X     & 0.13 & 363 & 41.53 & 0.09    \\ \cline{2-6} 
                           & A3C       & 1.70 & 368 & 51.07 & 0.08    \\ \hline
\multirow{2}{*}{beamrider} & APE-X     & 0.03 & 361 & 35.09 & 0.10    \\ \cline{2-6} 
                           & A3C       & 0.77 & 361 & 39.45 & 0.09    \\ \hline
\end{tabular}

\vspace*{0.5em}
{\small The third column shows the horizontal axis bandwidth obtained for each environment-algorithm combination (given in the first two columns), computed as described in \cref{eqn:bandwidth}, for $M = 760$, which is the size of the ``training set'' used in \cref{sec:statisticalvalidation}.  The fourth column gives the number of training data tuples that were generated via uniform sampling (see \cref{alg:evaluation}), and that were not filtered out when removing the $5\%$ data tuples with the highest proxy criticality values.  The fifth column shows the estimated sample size computed via \cref{eqn:samplesizeapproximation:gaussian}, but with $M$ set to the number given in the fourth column (e.g., $363$), rather than $760$, and with the result of the equation then divided by $2$.  Finally, the last column gives the estimated upper bound on the percentile error $\beta - \beta^{*}$, computed via \cref{eqn:percentilebound} based on the sample size $D$ in the fifth column, with a confidence level $\alpha = 0.95$, which has a corresponding $z_{\alpha} = 1.645$; we report the tighter upper bound (middle of \cref{eqn:percentilebound}).}
\end{table*}

In \cref{cai:tab:results:errors}, we estimate the sample sizes and bounds on percentile errors, using the approach presented above.  In order to come up with conservative estimates, we consider only those data tuples that were generated using uniform sampling, when setting the value of $M$ in \cref{eqn:samplesizeapproximation:gaussian}, since without uniform sampling, the proxy criticality values are highly non-uniform.  We also divide the result of the equation by $2$, to account for reduced data at interval boundaries $p_{min}(\cdot)$ and $p_{max}(\cdot)$, as suggested earlier.  For lower values of $p(\cdot)$, the estimates of $D$ are likely too low (and the estimates of $\epsilon_{percentile}$ bounds are thus probably too high), since many of the remaining data tuples (generated without uniform sampling) are clustered there.  We note that the numbers in \cref{cai:tab:results:errors} do not depend upon the number of time steps perturbed $n$, since the proxy criticality metric \cite{lin2017} does not depend on $n$, and thus, the bandwidth $H_{p(\cdot)}$ is independent of $n$ as well.

\subsubsection{Limitations}
While above analysis provides a very rough sense of the possible magnitude of the percentile error, it is not entirely reliable.  First, as we showed in \cref{fig:results:metrichistogram:beamrider} and \cref{fig:results:metrichistogram:pong}, the distribution of proxy criticality values is often non-uniform, resulting in different sample sizes and different bounds for different proxy criticality values/bins.  Furthermore, if $D$ is low for some bins, then the normal approximation of \cref{eqn:percentilebound} becomes less accurate.  Finally, within a given proxy criticality value $p(\cdot)$ value, the true criticality density is computed (via the use of kernels) from different data tuples, some of them having proxy criticality values higher than $p(\cdot)$, and others lower.  Since the true, population percentile is often dependent upon $p(\cdot)$ (typically, with a positive correlation between them), the proportion of density at or below $b_{\beta}\left(p(\cdot), n\right)$ no longer follows an exact binomial distribution; this can also make the bound in \cref{eqn:percentilebound} less accurate, especially, when bandwidth $H_{p(\cdot)}$ is high.  Because of these issues, rather than relying on \cref{eqn:percentilebound} and \cref{eqn:samplesizeapproximation:gaussian} to guarantee a bound on $\epsilon_{percentile}$ (given a sample size $M$), we instead empirically evaluated the quality of the percentiles empirically, through cross-validation.  That being said, the equations suggest that, at least for a sufficiently small $H_{p(\cdot)}$ and sufficiently large $M$, the bound on percentile error decreases with the square root of $M$, even when a KDE plot is used to compute percentiles.

\end{document}